\documentclass[letterpaper, 10 pt, conference]{ieeeconf}
\IEEEoverridecommandlockouts
\usepackage[pdftex]{graphicx}
\usepackage{array}
\usepackage{hyperref}
\usepackage[free-standing-units=true]{siunitx}
\usepackage[font=footnotesize]{subcaption}
\usepackage[font=footnotesize]{caption}

\usepackage[nobreak]{cite}
\usepackage[super]{nth}

\usepackage{algorithm}
\usepackage{algorithmic}

\usepackage{cancel} 
\usepackage{times} 
\usepackage{amssymb, amsmath} 

\usepackage{url}

\let\labelindent\relax 
\usepackage[inline]{enumitem}\setlist[enumerate]*{label=(\roman*)}

\usepackage{multirow}

\newcommand\figref[1]{\text{Fig.~\ref{#1}}}
\newcommand\tableref[1]{\text{Table~\ref{#1}}}

\newcommand\algorithmref[1]{\text{Alg.~\ref{#1}}}
\newcommand*{\sref}[1]{\S\ref{#1}}

\title{\LARGE \bf
KeyMPs: One-Shot Vision-Language Guided Motion Generation by Sequencing DMPs for Occlusion-Rich Tasks
}
\author{Edgar Anarossi$^{1}$, Yuhwan Kwon$^{1,2}$, Hirotaka Tahara$^{1,3}$, Shohei Tanaka$^{4}$, Keisuke Shirai$^{4}$, Masashi Hamaya$^{4}$, \\Cristian C. Beltran Hernandez$^{4}$, Atsushi Hashimoto$^{4}$, and Takamitsu Matsubara$^{1}$
\thanks{$^{1}$Authors are affiliated with the Division of Information Science, Graduate School of Science and Technology, Nara Institute of Science and Technology, Japan.}
\thanks{$^{2}$Author is affiliated with the Department of Electrical and Electronic Engineering, Faculty of Engineering Science, Kansai University, Osaka, Japan.}
\thanks{$^{3}$Author is affiliated with the Department of Electronics, Kobe City College of Technology, Hyogo, Japan.}
\thanks{$^{4}$Authors are affiliated with the OMRON SINIC X Corporation, Tokyo, Japan.}%
\thanks{This work was supported by JSPS KAKENHI Grant Numbers JP21H04910, JP24K03018.}%
}

\begin{document}
\maketitle
\thispagestyle{empty}
\pagestyle{empty}

\begin{abstract}
Dynamic Movement Primitives (DMPs) provide a flexible framework wherein smooth robotic motions are encoded into modular parameters. 
However, they face challenges in integrating multimodal inputs commonly used in robotics like vision and language into their framework.
To fully maximize DMPs' potential, enabling them to handle multimodal inputs is essential.
In addition, we also aim to extend DMPs' capability to handle object-focused tasks requiring one-shot complex motion generation, as observation occlusion could easily happen mid-execution in such tasks (e.g., knife occlusion in cake icing, hand occlusion in dough kneading, etc.).
A promising approach is to leverage Vision-Language Models (VLMs), which process multimodal data and can grasp high-level concepts.
However, they typically lack enough knowledge and capabilities to directly infer low-level motion details and instead only serve as a bridge between high-level instructions and low-level control. 
To address this limitation, we propose Keyword Labeled Primitive Selection and Keypoint Pairs Generation Guided Movement Primitives (KeyMPs), a framework that combines VLMs with sequencing of DMPs.
KeyMPs use VLMs' high-level reasoning capability to select a reference primitive through \emph{keyword labeled primitive selection} and VLMs' spatial awareness to generate spatial scaling parameters used for sequencing DMPs by generalizing the overall motion through \emph{keypoint pairs generation}, which together enable one-shot vision-language guided motion generation that aligns with the intent expressed in the multimodal input.
We validate our approach through experiments on two occlusion-rich tasks: object cutting, conducted in both simulated and real-world environments, and cake icing, performed in simulation.
These evaluations demonstrate superior performance over other DMP-based methods that integrate VLM support.
\end{abstract}

\section{INTRODUCTION}

\begin{figure}
\centering
\includegraphics[width=1.0\hsize] {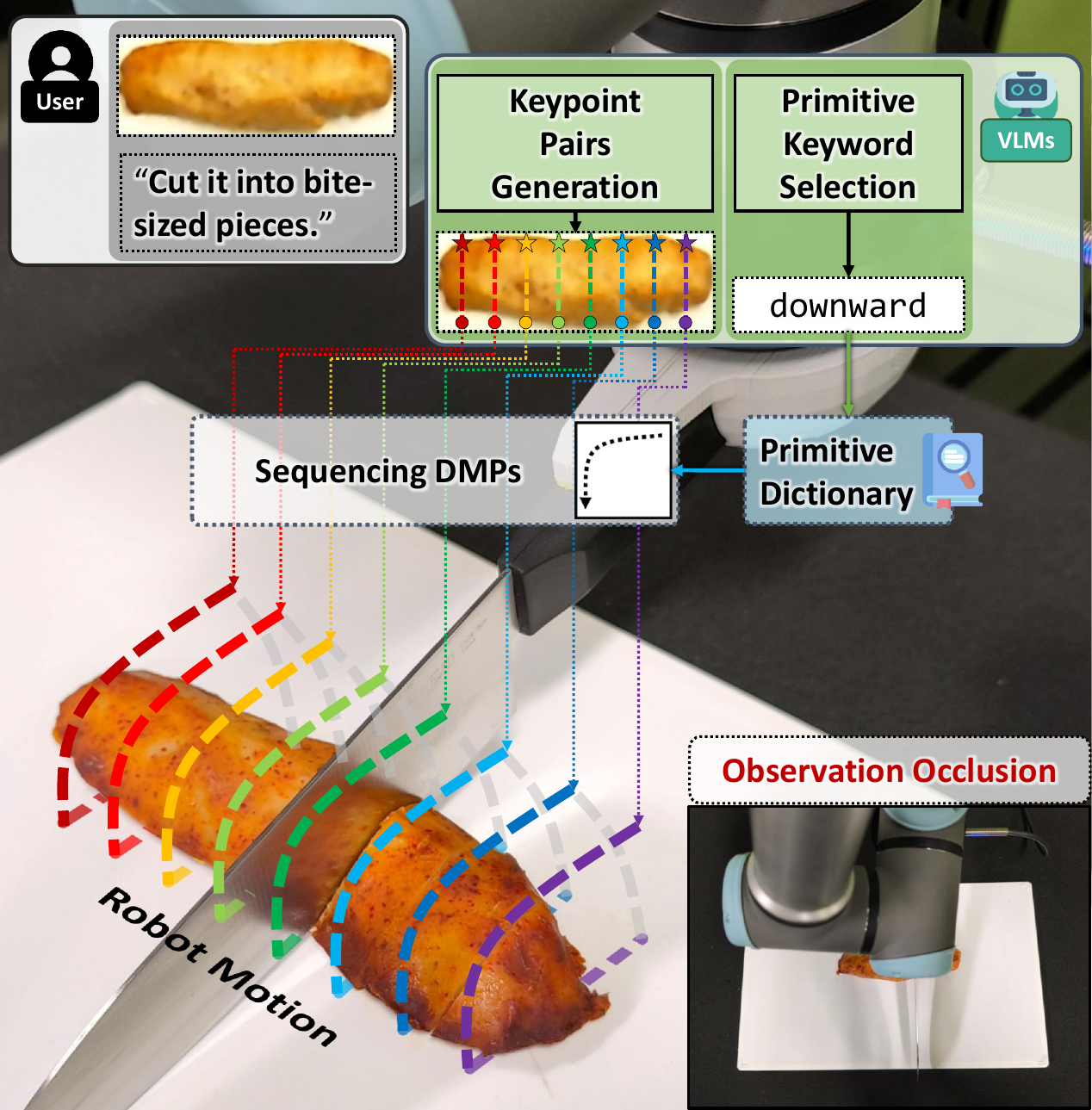}
\caption{
Execution of cutting motion generated by KeyMPs. 
The framework leverages Vision-Language Models (VLMs) to select Dynamic Movement Primitives (DMPs) learned parameters and generate keypoint pairs for sequencing DMPs according to the user's intention.
}
\label{fig:1}
\end{figure}

\PARstart{D}{ynamic} Movement Primitives (DMPs) are a powerful framework for robotic motion generation. 
They encode motions compactly with stability and robustness due to their foundation in dynamical systems \cite{ijspeert2013dynamical, schaal2007dynamics, ude2010task}.
DMPs enable efficient learning and reproduction of motor behaviors and offer scalability in spatial and temporal domains, which are crucial for performing diverse tasks in different environments \cite{pastor2009learning, kober2008policy, stulp2011hierarchical}. 
The modularity and parameterization of DMPs generate motion in a flexible manner wherein motions are encoded into easily integrated parameters that enable them to adapt to different environmental conditions \cite{paraschos2013probabilistic, calinon2016tutorial}. However, DMPs have an inability to handle effectively multimodal inputs like vision and language data that are commonly used in robotics.

Building upon DMPs' strengths, we aim to improve their flexibility by incorporating vision and language inputs, thereby expanding robots’ ability to interact with humans and environments using natural communication and perception \cite{stepputtis2020language}.
In particular, we seek to enable one-shot vision-language guided motion generation since continuous observation might not be feasible in occlusion-rich tasks such as would be encountered in the kitchen, i.e., food cutting, cake icing, dough kneading, etc.

In order to achieve this goal, two main objectives must be met, i.e., associating DMPs with language and vision inputs\cite{tellex2011understanding, ahn2022can, driess2023palm} and extending DMPs to tasks requiring one-shot vision-language guided complex motion generation, as current DMP-based motion generation methods \cite{pahic2018deep, pahic2020training} focus on relatively simple motions.
Achieving these objectives requires overcoming challenges, particularly those from deep-learning methods.
In particular, associating DMP parameters with high-dimensional data requires extensive datasets for generalization, which thus limits their real-world use \cite{pahic2018deep, pahic2020training}
Additionally, interpreting linguistic inputs \cite{shridhar2022cliport, stepputtis2020language} and generating long, complex DMP motions in one-shot \cite{pahic2020training} expand the required feature space. 
This issue intensifies the need for even larger datasets beyond what is already required for high-dimensional input. 
Addressing these challenges is crucial for executing intricate tasks based on high-level instructions and visual cues.

To address these challenges, we integrate Vision-Language Models (VLMs) with sequencing of DMPs to interpret and generate complex motions from vision and language input. 
VLMs have been shown to effectively bridge language instructions and visual observations, enabling more accurate task planning \cite{shirai2024vision} and execution \cite{10870896}.
However, it is generally ineffective to generate DMP parameters directly from VLMs, since these models are not trained to handle such low-level motion representations.
As a solution, we tried to leverage the key features of VLMs, which are their effective natural language processing that provides reasoning capabilities through conceptual understanding and their spatial awareness, in generating task-relevant keypoints to be used in the overall motion planning.

Building on these features, we propose \textbf{Keyword Labeled Primitive Selection and Keypoint Pairs Generation Guided Movement Primitives (KeyMPs)}, a framework that integrates VLMs into DMP-based motion generation. 
The framework leverages VLMs' strengths through two main components.
The first, called Keyword Labeled Primitive Selection, uses VLMs' natural language processing to select task-related intuitive labels mapped to DMP parameters (referred to as \emph{learned parameters}) obtained via imitation learning \cite{zhou2023generalizable, liu2024enhancing}, thereby bridging natural language with DMPs. 
The second, called Keypoint Pairs Generation, exploits VLMs' spatial awareness to generate multiple keypoint pairs. Each generated keypoint pair serves as distinct \emph{spatial scaling parameters} which will be used to facilitate sequencing DMPs.
We enhance VLMs’ 3D capabilities \cite{chen2024spatialvlm} by integrating real-world depth cues, such as an object’s or obstacle's height. 
This allows a simplified 2D keypoint representation for spatial parameters while preserving key 3D details.

The \textbf{KeyMPs} framework involves creating a dictionary of DMP parameters learned from demonstrated motions, labeling each primitive, and using VLM-based components to interpret vision-language guides.
During the execution phase, to avoid observation occlusion mid-execution (see \figref{fig:1}), visual observation and desired outcome described in text are acquired before execution.
These inputs guide the Primitive Keyword Selection to select the proper primitive from the reference primitive dictionary.
Concurrently, the Keypoint Pairs Generation translates visual inputs into the desired 2D keypoint pairs design by transforming them into spatial scaling parameters for sequencing multiple DMPs.
Finally, the robot motion is generated in one-shot through sequencing DMP motions reconstructed from the learned parameters scaled by the spatial scaling parameters.

The key contributions of this paper are:
\begin{enumerate}
    \item Introducing the \textbf{KeyMPs} framework, a novel framework that performs sequencing DMPs from visual and language inputs in one-shot by utilizing VLMs to handle primitive selection and keypoint pairs generation.
    \item Demonstrating \textbf{KeyMPs}’ effectiveness in simulated environments through a comprehensive analysis of an occlusion-rich tasks, executable 3D motions that align with the intent expressed in the multimodal input, and showcasing its superior task generalization compared to deep learning-based and ablation methods.
    \item Validating and demonstrating the effectiveness of \textbf{KeyMPs} through experiments involving a real robot performing an occlusion-rich task in a real environment with real-world images, as well as validating the executability of the generated motion with a real robot.
\end{enumerate}

\section{RELATED WORK}

\subsection{Deep Learning-based DMPs Frameworks with Vision Input}

The integration of DMPs with deep learning has expanded their applicability to robotics by enabling more flexible and adaptive motion generation \cite{ijspeert2013dynamical}. 
Various deep-imitation learning frameworks, such as Convolutional Image-to-Motion Encoder-Decoder Network (CIMEDNet) \cite{pahic2018deep, pahic2020training} and Deep Segmented DMPs Network (DSDNet) \cite{anarossi2023deep}, estimate the parameters of DMPs from visual data, allowing robots to mimic detailed motions.
Similarly, certain deep reinforcement learning approaches incorporate DMPs as structured policies, which are then optimized through environment interactions \cite{stulp2012reinforcement}.

However, combining deep learning with DMPs presents challenges, particularly in tasks requiring a high level of generalization. 
Existing methods which attempt to associate DMPs with high-dimensional input data often require large datasets for proper generalization \cite{pahic2018deep, pahic2020training}, limiting real-world applications due to the high costs and difficulties of data acquisition and annotation. 
Incorporating both vision and language increases these demands, often necessitating hundreds of demonstrations and thousands of simulated environments for effective performance \cite{stepputtis2020language}.
Additionally, extending DMPs in a way that they can generate long, complex motions also leads to computational inefficiencies in the form of multiple acceleration phases and high-dimensional feature spaces \cite{pahic2020training}. 

Among these deep-learning-based DMP frameworks, only CIMEDNet \cite{pahic2020training}, which generates motion by using a set of DMP parameters, and our prior work on DSDNet \cite{anarossi2023deep}, which sequences DMPs by generating multiple sets of DMP parameters, are capable of planning complete task-oriented motion sequences from an image input.
Despite this capability, the performance of both models remains severely limited by a lack of training data, which hinders their adaptability to new situations and tasks.
To address these limitations, we used cross-domain knowledge through VLMs in this research to provide an understanding of visual input and essential insights to develop intricate robotic motions without requiring extra fine-tuning and that adapt to new situations.

\subsection{LLM and VLM Integrated DMP Frameworks}

Large Language Models (LLMs), such as GPT, PaLM, and other large language models \cite{brown2020language, devlin2018bert, touvron2023llama2, chowdhery2023palm} have significantly advanced robotics by enhancing human-robot interaction and decision-making through language understanding. 
These models interpret language commands and generate corresponding actions \cite{vemprala2024chatgpt, ahn2022can}. 
VLMs further integrate visual perception with language comprehension, enabling robots to act on multimodal data and perform tasks requiring both vision and language \cite{alayrac2022flamingo, li2022blip, brohan2022rt, driess2023palm}.
The integration of these foundation models into robotics systems allows conventional methods to be more flexible and scalable \cite{kawaharazuka2024real}.

Recent research has explored combining LLMs with DMPs in order to enhance robotic motion generation \cite{liu2024enhancing, zhou2023generalizable}. 
It has shown that high-level language instructions can be translated into low-level motion primitives for complex task planning.
However, as tasks become increasingly detailed or require contextual information from visual inputs, relying solely on language-driven approaches can make it challenging to generate precise positional data \cite{brohan2022rt} or manage different groups of DMP parameters \cite{liu2024enhancing, zhou2023generalizable}.

Despite advancements, the methods discussed above require continuous feedback to VLMs and generate only short motion segments rather than complete motion sequences \cite{liu2024enhancing, zhou2023generalizable}, which are impractical for occlusion-rich tasks. 
To address this limitation, our framework leverages the VLMs' spatial awareness to plan the overall motion, thereby enabling one-shot motion planning for complex tasks without the need for constant feedback.

\subsection{Object Cutting in Robotics}

The field of robotic object cutting has progressed through the development of diverse methodologies aimed at enabling robots to perform precise and adaptive cutting tasks \cite{mu2019robotic}. 
Some approaches emphasize dynamic force control utilizing sensor feedback to regulate knife motion during slicing \cite{mu2019robotic}, while others integrate cutting into broader task planning frameworks, sequencing actions derived from, e.g., cooking recipes \cite{inagawa2021analysis,schmitz2023robot}. 
Additionally, machine-learning-based methods have used simulations or real-world data to train policies for cutting multi-material objects \cite{xu2023roboninja,beltran2024sliceit}. 
These frameworks showcase a range of strategies, from low-level control to high-level planning, and often rely on training data or predefined models to achieve their objectives.

However, generating complex cutting trajectories remains a significant challenge for these methods, particularly in terms of generalization and flexibility. 
Many frameworks depend heavily on extensive datasets or calibrated simulations \cite{schmitz2023robot,beltran2024sliceit,xu2023roboninja}, which restrict their practical applicability due to the costs of data acquisition and computation. 
Furthermore, these approaches often struggle to adapt to unseen objects or intricate cutting sequences without predefined trajectories or motion models \cite{inagawa2021analysis,schmitz2023robot}. 
In this research, we address these limitations by focusing on cutting-trajectory generation and leveraging cross-domain knowledge through VLMs to interpret visual and linguistic inputs and produce intricate cutting trajectories without requiring extensive training data.

\section{PRELIMINARY}

\subsection{Single-DMP Formulation}

DMPs have long served as a foundational framework for representing and executing robotic motion \cite{ijspeert2013dynamical}. 
A single DMP is commonly described by the following set of differential equations:
\begin{align}
    \tau \dot{z}(t) &= \alpha_z \Bigl(\beta_z \bigl(y_{\text{goal}} - y(t)\bigr) - z(t)\Bigr) + f\bigl(s(t)\bigr), 
        \label{eq:dmp_acceleration}\\
    \tau \dot{y}(t) &= z(t), 
        \label{eq:dmp_velocity}\\
    \tau \dot{s}(t) &= -\alpha_s\,s(t). 
        \label{eq:dmp_canonical}
\end{align}
Here, \( y(t) \) represents the system's position at time \( t \), which dynamically evolves towards the goal position \( y_{\text{goal}} \) under the influence of attractor dynamics. 
The scaled velocity \( z(t) \) dictates the rate of motion, while the phase variable \( s(t) \) decays over time to ensure a smooth progression through the motion. 
The temporal scaling factor \( \tau \) adjusts the execution speed, and the constants \( \alpha_z \), \( \beta_z \), and \( \alpha_s \) govern the system's stability and convergence behavior. 
The forcing function \( f(s(t)) \) introduces non-linearities, which enable the DMP to generate complex trajectories beyond simple point-to-point motions:
\begin{align}
    f\bigl(s(t)\bigr) &= 
        \sum_{i=1}^{N} w_i\,\psi_i\bigl(s(t)\bigr)\,s(t), \\
    \quad\text{with}\quad
    \psi_i\bigl(s(t)\bigr) &= 
        \exp\!\bigl(-h_i\,[s(t)-c_i\,]^2\bigr).
    \label{eq:dmp_forcing_function}
\end{align}
In these equations, \( N \) denotes the number of basis functions, each \( \psi \) is typically a Gaussian function with center \( c \) and width \( h \), and \( w \) are learnable weights derived from demonstrations. 
\(w\) denotes the DMPs' learned parameters, while \( \{y_0, y_{\text{goal}}\} \) are the DMPs' spatial scaling parameters that define the starting and goal positions in the task space.

\subsection{Sequencing Multiple DMPs in Time}

\begin{figure*}[htp]
\centering
\includegraphics[width=1.0\hsize]{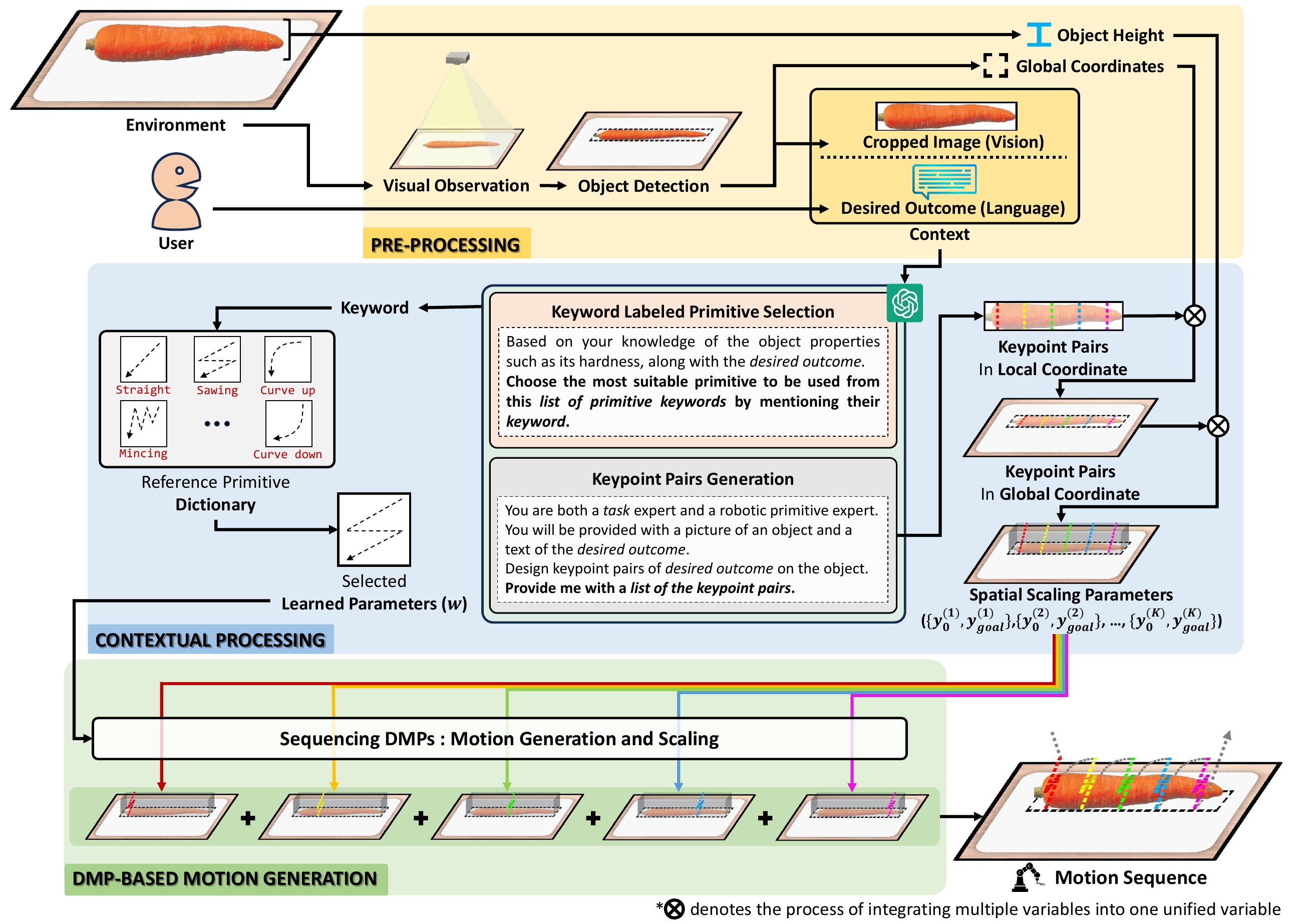}
\caption{
Overview of the KeyMPs framework, illustrated with an example task of object cutting. 
The framework processes inputs consisting of an RGB image and natural language text.
Object detection identifies the global position of the object, and the image is cropped accordingly. 
The cropped image and text are then processed using VLM-based components. 
The Keyword Labeled Primitive Selection part selects DMPs' learned parameters, and Keypoint Pairs Generation part creates the base for the scaling parameters. 
These 2D keypoint pairs are augmented with the global position and object's height to generate 3D spatial scaling parameters that scale the primitives, which are subsequently sequenced to produce the final executable motion.
} 
\label{fig:keydmps}
\end{figure*}

For more intricate behaviors, multiple DMPs can be sequenced in time \cite{saveriano2023dynamic,manschitz2014learning, cho2020learning, li2017reinforcement}, with each handling a segment of the motion. 
Here, let \(K\) be the total number of segments, with time boundaries $t_0 \;<\; t_1 \;<\;\cdots\;<\; t_K,$, where \(t_0\) and \(t_K\) are the start and end times of the entire motion.
The overall trajectory \(\mathbf{Y}(t)\) is defined as a piecewise function:
\begin{equation}
    \mathbf{Y}(t)
    \;=\;
    \begin{cases}
        \mathbf{Y}_1(t), 
            & t_0 \;\le\; t \;<\; t_1,\\[5pt]
        \mathbf{Y}_2(t), 
            & t_1 \;\le\; t \;<\; t_2,\\[-1pt]
        \multicolumn{2}{c}{\vdots} \\
        \mathbf{Y}_K(t), 
            & t_{K-1} \;\le\; t \;\le\; t_K,
    \end{cases}
    \label{eq:dmp_concat}
\end{equation}
where each sub-trajectory \(\mathbf{Y}_k(t)\) is generated by a separate DMP, typically with its own parameters $\bigl\{\tau^{(k)}, \alpha_z^{(k)}, \beta_z^{(k)}, w_i^{(k)}, \\ y_0^{(k)}, y_{\text{goal}}^{(k)}\bigr\}$, and is integrated over the interval \([t_{k-1},\,t_k)\). 

A piecewise time-based definition of \(\mathbf{Y}(t)\) as shown in \eqref{eq:dmp_concat} is often used in practice to compose complex behaviors from multiple DMPs, each focusing on a simpler sub-motion (e.g., approach, cut, retreat). This modularity can increase data efficiency and adaptability while preserving a clear mapping to real-time robotic control loops. Depending on application requirements, segments may be purely time-based or event-based (e.g., a segment ends when a sensor detects contact).

\section{PROPOSED FRAMEWORK}
\label{proposed-framework}

Here, we present our framework that leverages pre-trained VLMs for one-shot vision-language guided motion generation through sequencing DMPs.
First, we provide an overview of the framework. 
Then, we describe the input pre-processing, keyword labeled primitive selection, keypoint pairs generation and transformation, and construction of the generated motion sequence using DMPs.

\subsection{Framework Overview}
\label{method-overview}

Our framework, \textbf{KeyMPs} (as shown in \figref{fig:keydmps}), integrates vision and language inputs to generate executable motions by leveraging VLMs and sequencing DMPs. 
It operates in three stages: 
\begin{enumerate}
    \item \textbf{Pre-Processing:} Collects the necessary inputs, including language instructions, visual observations, and object-specific information such as the object's height, for processing by the VLM-based components described in \sref{primitive-selection} and \sref{pattern-generation}.
    \item \textbf{Contextual processing:} Takes the vision and language context from the pre-processing stage and performs structured component decomposition by separately processing 
    \begin{enumerate*}
    \item selection of learned parameters by using keyword labeled primitive selection and 
    \item generation of spatial scaling parameters using keypoint pairs generation.
    \end{enumerate*}
    \item \textbf{DMP-based motion generation:} Combines the learned parameters within the selected primitive with the generated spatial scaling parameters from the generated keypoint pairs through sequencing DMPs in order to create the robot motion.
\end{enumerate}

\subsection{Pre-Processing}
\label{input-acquisition}

Our framework relies on two primary types of input: visual and textual. 
The raw visual input is captured through a camera as an environment observation image. 
An object detector is then used to extract the object's global coordinates and crop the image to focus on the object of interest. 
This transformation can be expressed as:
\begin{align}
    pos_{global},\,img_{obj} &= \mathrm{ObjectDetector}(img_{env}),
    \label{eq:obj_detector}
\end{align}
where \(img_{env}\) is the raw environment observation image, \(pos_{global}\) represents the object's global coordinates as determined by the object detector, and \(img_{obj}\) is the resulting cropped image.

In addition to visual input, the framework accepts textual input $l$ that provides a more detailed task description on what the user desires in natural language, complementing the general task initialized in the VLMs. 
Together, these inputs form the foundation for the VLM-based components. 
The framework also requires object-specific information, in particular, the object's height \(h\), which is integrated during post-processing to generate spatial scaling parameters. 
Various acquisition methods can be used to obtain \(h\); this offers flexibility in choosing the sensor or measurement technique that suits the application.

\begin{figure}[t]
\includegraphics[width=1.0\hsize]{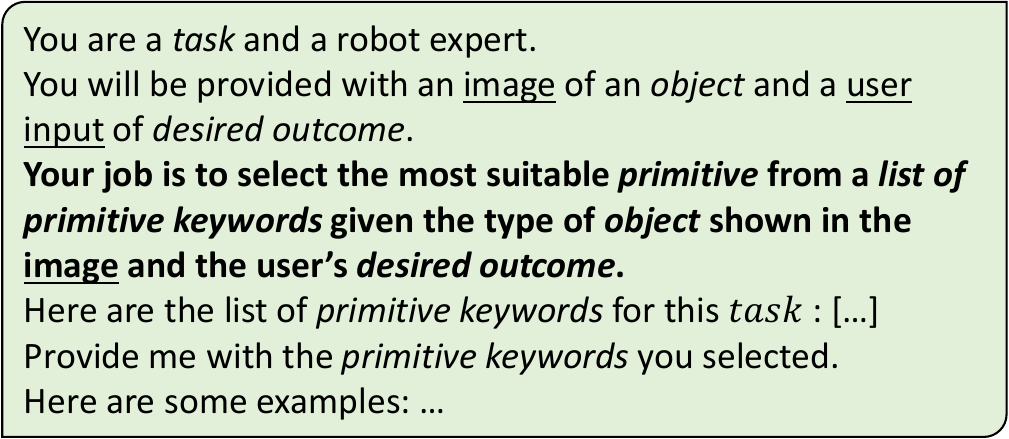}
\caption{
Keyword labeled primitive selection initialization prompt.
}
\label{prompt:primitive-selection}
\end{figure}

\subsection{Contextual Processing}

\subsubsection{Keyword Labeled Primitive Selection}
\label{primitive-selection}

The application, the type of primitive that is used in the task is an important choice that depends on the application. 
For example, in a food-cutting task, the outcome may differ significantly depending on the style of cutting motion employed.
To account for these variations, we assume to have access to a dictionary of DMPs' learned parameters $w$, with each representing a different action/motion style as shown in \figref{fig:keydmps}.
We employ VLMs with advanced reasoning capabilities to appropriately map the user instruction and environment observation with the available primitives.

A primitive dictionary \(D\) is created to map each descriptive \( keyword \) to its corresponding learned basis function weights \(w\).
To achieve this, a VLM-based component that processes an image and accompanying textual input to output the appropriate keyword is utilized.
The process is expressed as:
\begin{align}
    keyword &= \mathrm{VLM_{keyword}}(img_{obj},\,{l}), \\
    w &= D(keyword).
    \label{eq:VLM_mapping}
\end{align}
Here, \( \mathrm{VLM_{keyword}}\) is a VLMs initialized by a system prompt described in \figref{prompt:primitive-selection}, where the details of the current task being handled, a list of primitives that can be used, and several task-related examples are provided to guide the selection of appropriate primitives.
On execution, \( \mathrm{VLM_{keyword}}\) takes the cropped image of the object $img_{obj}$ and the natural language input $l$ to produce a descriptive $keyword$.
The dictionary D is then used to map this keyword to the corresponding DMP learned parameters \(w\) of the selected primitive.

This approach requires collecting a small dataset of basis function weights for the specific primitives needed for the task. 
Since there are no universal rules for selecting primitives across all tasks, we therefore rely on domain knowledge to guide the selection process. 
For example, in the object cutting task, the required cutting primitives often vary depending on the object's properties such as hardness or texture. 
Harder or crusty objects may require a sawing primitive, while softer objects can be cut using a downward primitive. 
Ultimately, the selection of appropriate primitives depends on the specific task at hand.

\begin{figure}[t]
\centering
\includegraphics[width=1.0\hsize]{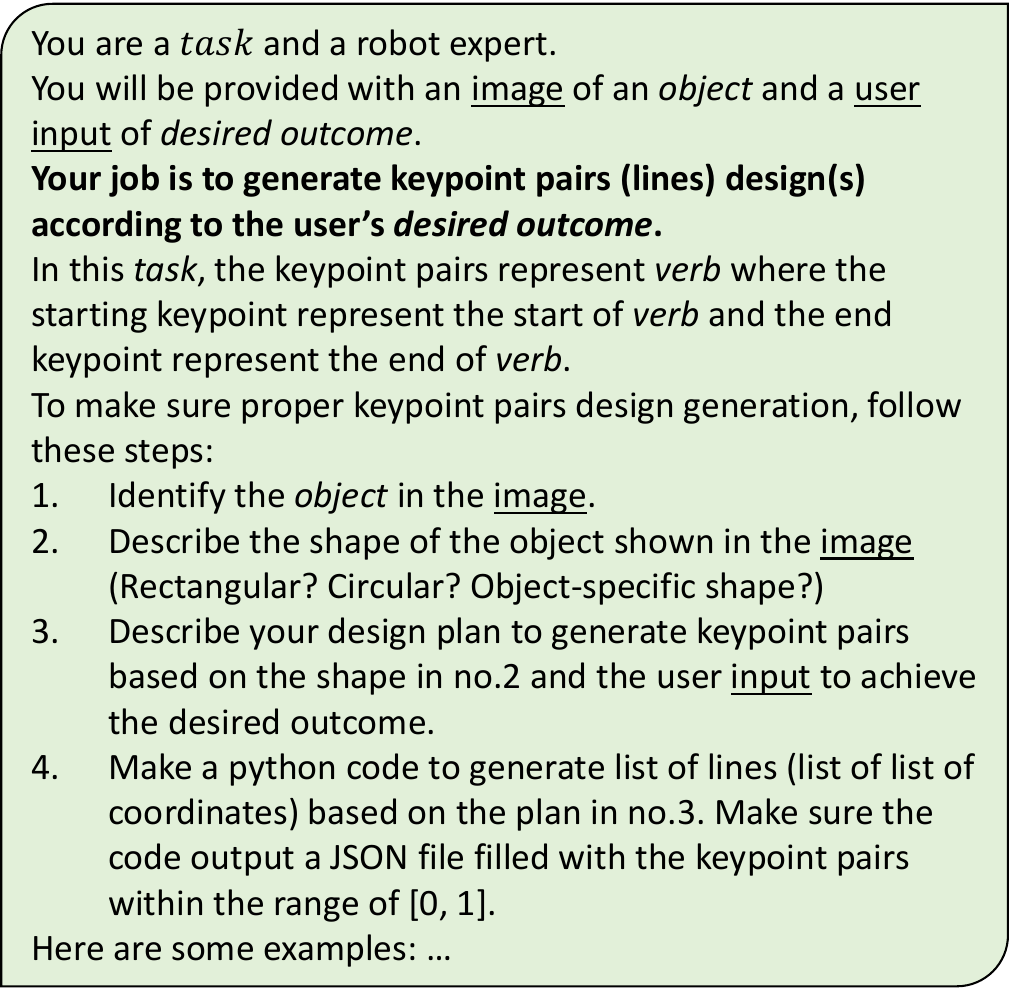}
\caption{
Keypoint pairs generation initialization prompt.
}
\label{prompt:keypoint-generation}

\end{figure}

\subsubsection{Keypoint Pairs Generation}
\label{pattern-generation}

To generate spatial scaling parameters for the DMPs, we leverage VLMs' ability to comprehend spatial coordinates on the basis of language and visual inputs. 
Specifically, a VLM-based component is utilized to produce $K$ 2D keypoint pairs in pixel-space, effectively representing line segments used as the base value of $K$ start and goal positions for each DMPs to be sequenced. 
We express this process as follows:
\begin{align}
    keypoint\ pairs &= \mathrm{VLM_{keypoints}}(img_{obj},\,l), \\
    \mathbf{y_0,\,y_{goal}} &= \mathrm{PostProcess}(keypoint\ pairs,  pos_{global}, h).
    \label{eq:vlm_keypoints}
 \end{align}
Here, \(\mathrm{VLM_{keypoints}}\) is another VLMs initialized by a different system prompt shown in \figref{prompt:keypoint-generation} where the details of the current task being handled, how the VLMs are supposed to generate the keypoint pairs, and several simple descriptive examples are provided.
This prompt guides the VLMs to translate the desired outcome of varying specificity into a number of keypoint pairs, $K$, which is not pre-specified and is instead determined by the VLMs based on the given context. 
These keypoint pairs can be visually verified by projecting them onto the image before motion generation.
On execution, \(\mathrm{VLM_{keypoints}}\) takes the same cropped image of the object $img_{obj}$ and the natural language input $l$ to generate $K$ 2D $keypoint\ pairs$ (lines) in the pixel-space.

Following generation of these keypoint pairs, \(\mathrm{PostProcess}\) applies additional transformations, beginning with a 2D transformation to map the pairs into the object's global coordinates (Appendix B), followed by integration of height information, which can differ from one task to another (Appendix C). 
These steps, encompassing global coordinate conversion and height integration, finalize the 3D context and yield $K$ modified keypoint pairs as valid DMP spatial scaling parameters \(\bigl(y_0,\, y_{\text{goal}}\bigr)\).

\subsection{DMP-based Motion Generation}
\label{motion-generation}

In this stage, we construct the DMP-based motion by sequencing multiple DMP instances.
For each keypoint pair from \sref{pattern-generation}, separate DMPs are instantiated by scaling the learned parameters of the primitive selected in \sref{primitive-selection} with the corresponding spatial scaling parameters.
The overall flow of our framework from pre-processing to motion generation is presented in \algorithmref{algorithm:motion-generation}.

For each of the keypoint pair:
\begin{enumerate}
    \item Append to the motion sequence a DMP-based translation motion to move the robot's end effector from its current position to the starting position of the keypoint pair.
    \item Scale the reference primitive by adjusting its initial position ($y_0$) and goal position ($y_{\text{goal}}$) to match the new positions provided by the keypoint pair.
    \item Append to the motion sequence the DMPs motion scaled reference primitive.
\end{enumerate}

\begin{algorithm}[t]
\caption{\footnotesize{KeyMPs Motion Generation}}
\label{algorithm:motion-generation}
\begin{algorithmic}[1]
\STATE \textbf{Parameters:}
\STATE \quad $img_{env}$ - Environment Visual Input
\STATE \quad $l$ - Natural Language Input
\STATE \quad $h$ - Object's Height
\STATE \textbf{Initialize:}
\STATE \quad $D$ - Reference Primitive Dictionary
\STATE \quad $pos_{global},img_{obj} \gets \mathrm{ObjectDetector}(img_{env})$
\vspace{0.5em}
\STATE $keyword \gets \mathrm{VLM_{keyword}}(img_{obj},l)$
\STATE $keypoint\ pairs \gets \mathrm{VLM_{keypoints}}(img_{obj},l,pos_{global},h)$
\STATE $motion\ sequence \gets \mathrm{DMPMotionGen}(D,keyword,$ \\ $keypoint\ pairs)$
\RETURN $motion\ sequence$
\end{algorithmic}
\end{algorithm}

\section{SIMULATION EXPERIMENT}
\label{sim-exp}

\begin{figure}[t]

    \begin{minipage}[b]{1\linewidth}
    \centering
    \includegraphics[width=1.0\hsize]{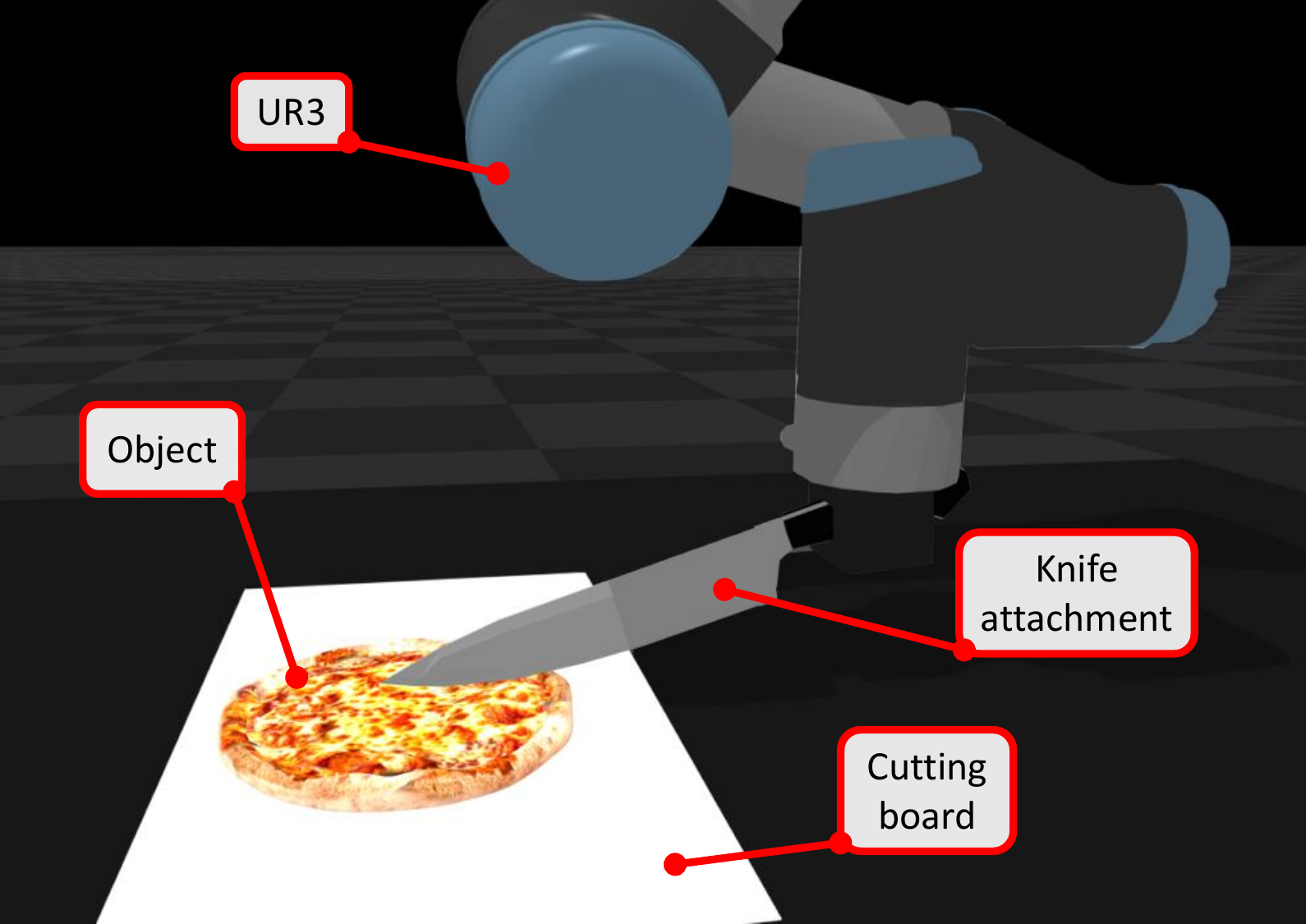}
    \small (a)
    \vspace{3truemm}
    \end{minipage}
    \begin{minipage}[b]{1\linewidth}
    \centering
    \includegraphics[width=1.0\hsize]{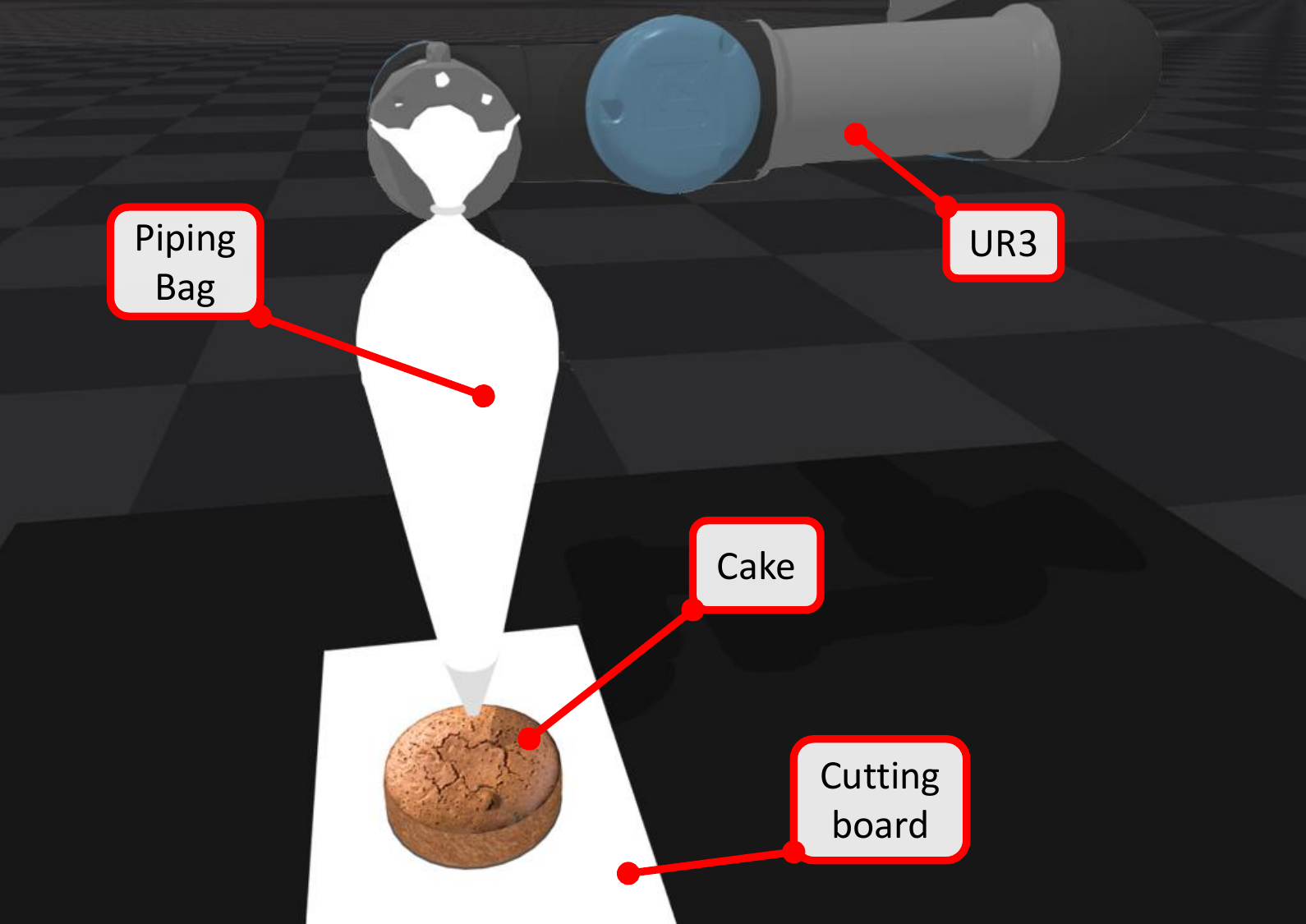}
    \small (b)
    \end{minipage}
    
\caption{Simulation environments in Isaac Gym for (a) the object cutting task and (b) the cake icing task. In both scenarios, the scene is observed by a single top-down RGB camera. During execution, the robot's tool obstructs the view of the object, which makes a one-shot motion generation approach essential.}
\label{fig:sim_env}
\end{figure}

To evaluate the effectiveness of our \textbf{KeyMPs} framework and address key research questions, we selected two occlusion-rich tasks: object cutting task and cake icing task. 
Both tasks require one-shot complex motion generation because the knife or piping bag would occlude visual observations during execution.
We created an environment in Isaac Gym \cite{makoviychuk2021isaac} for both tasks, as shown in \figref{fig:sim_env}(a) for the object cutting task and \figref{fig:sim_env}(b) for the cake icing task. We conducted four experiments designed to test various aspects of our framework using these two tasks.

\subsection{Research Questions}

We aim to address the following key questions:

\begin{itemize}
    \item[(P1)] Can VLMs generate an executable motion that aligns with the intent expressed in the vision-language input without structured component decomposition? (\sref{sim-exp2})
    \item[(P2)] How much does each VLM-based component in our framework contribute to the overall motion sequence? (\sref{sim-exp3})
    \item[(P3)] Does our framework achieve better generalization to unseen tasks than deep learning-based approaches \cite{pahic2020training, anarossi2023deep}? (\sref{sim-exp1})
    \item[(P4)] Can the proposed method demonstrate generalizability to tasks other than cutting? (\sref{sim-exp4})
\end{itemize}

In the first experiment, we compared methods that directly generate DMP-based motion using VLMs to address (P1).
For the second experiment, we checked the contribution of each VLM-based component separately.
Finally, for the third experiment, we tested the generalization performance of our method by comparing it with existing deep-learning-based DMP frameworks.
Finally, for the fourth experiment, we assessed the VLMs' ability to generate complex keypoint pairs for cake icing tasks.

\subsection{Experimental Setting}

We performed three experiments focusing on the object cutting task to answer our research questions.
Below, we outline the task details that serve as the foundation for our investigation.

\subsubsection{Object Cutting Task}
\label{sim-exp-cut}

\paragraph{Task Description}

The goal of this task is to cut objects on a cutting board measuring \SI{0.18}{\meter} by \SI{0.30}{\meter}. 
The task involves cutting objects of varying sizes and characteristics, which require different numbers and types of cuts. 
Rather than focusing on the physical interaction between the knife and the object \cite{beltran2024sliceit}, we emphasize the variety of cutting primitives and the cutting designs used to satisfy the user's intent.

To quantitatively evaluate performance in all simulation experiments, we designed various task scenarios detailed in \tableref{table:cutting-tasks}. 
These scenarios are sufficiently specific to yield nearly unique solutions, which ensures a robust assessment of the system's capability of handling diverse cutting requirements.

\begin{table}[t]
\caption{\textbf{List of prepared cutting tasks for simulation experiments}}
\label{table}
\centering
\setlength{\tabcolsep}{3pt}
\begin{tabular}{|c|l|c|>{\arraybackslash}p{50truemm}|}
\hline
 & Case & Object & \multicolumn{1}{c|}{Input Prompt} \\
\hline
\multirow{16}{*}{\rotatebox{90}{Trained Task}}  & 1 & Cabbage      & A single horizontal slice in the middle \\
                                                & 2 & Banana Bread & I want to eat 1 slice for each day of this week, cut it vertically \\
                                                & 3 & Round cake    & I'm having a party for 10 people, cut 1 slice for each \\
                                                & 4 & Round Pizza  & Cut it into 8 equal slices \\
                                                & 5 & Eggplant     & The object is 10 cm long, cut it vertically into 5 cm slices \\
                                                & 6 & Eggplant     & The object is 15 cm long, cut it vertically into 5 cm slices \\
                                                & 7 & Eggplant     & The object is 20 cm long, cut it vertically into 5 cm slices \\
                                                & 8 & Eggplant     & The object is 25 cm long, cut it vertically into 5 cm slices \\
                                                & 9 & Eggplant     & The object is 30 cm long, cut it vertically into 5 cm slices \\
\hline
\multirow{10}{*}{\rotatebox{90}{Unseen Task}}   & 10 & Cabbage & Slice the object into 3 parts horizontally \\
                                                & 11 & Eggplant & Slice both tips of the object  \\
                                                & 12 & Eggplant & The object is 35 cm long, cut it vertically into 5 cm slices  \\
                                                & 13 & Eggplant & The object is 40 cm long, cut it vertically into 5 cm slices  \\
                                                & 14 & Baguette & The object is 40 cm long, cut it vertically into 5 cm slices  \\
                                                & 15 & Baguette & The object is 45 cm long, cut it vertically into 5 cm slices  \\
\hline
\end{tabular}
\label{table:cutting-tasks}
\end{table}

\paragraph{Evaluation Method}
\label{task-evaluation}

For each scenario in \tableref{table:cutting-tasks}, we first established ground-truth trajectories by applying kitchen-based domain knowledge to convert textual commands into precise geometric paths, such as by measuring an object's dimensions to calculate a series of evenly spaced cuts. 
To validate that these patterns represent a common consensus, we subsequently conducted a user survey with 5 subjects (see Appendix E). 
The survey confirmed that our predefined lines aligned with the most widely agreed-upon cutting strategies for the given tasks. 
These validated 2D patterns were then systematically converted into 3D ground-truth coordinates by scaling the relevant primitive with the object's height.

These 3D coordinates served two purposes. First, they were the benchmark for our quantitative evaluation. 
Second, for Experiment 3, they were used to generate the training data for the deep-learning baselines through imitation.
To ensure the evaluation captured robustness, these ground-truth motions were also applied to objects with slightly randomized dimensions, accounting for subtle variations in object size.

\begin{figure}[t]
    \centering
    \includegraphics[width=\linewidth]{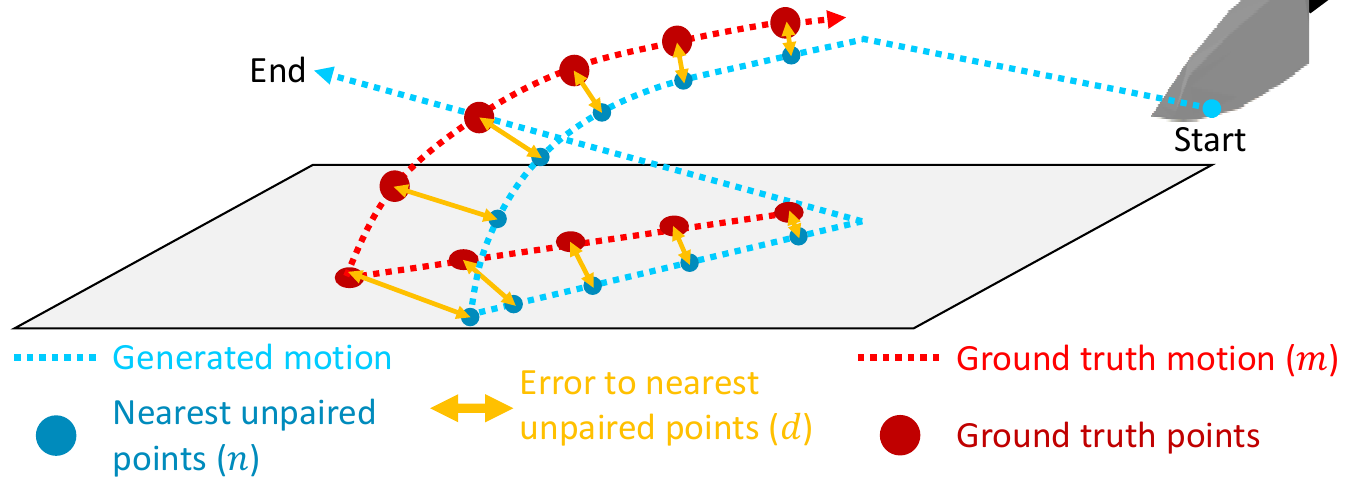}
    \caption{Evaluation visualization comparing generated motion (light blue) to ground-truth motion (red). Each unpaired ground-truth point (crimson) is matched to the closest unpaired point in the generated motion (blue). This metric addresses variability in the number of points and focuses the comparison on the significant parts of the trajectory, excluding preparation or finishing motions.}
    \label{fig:closest_point_visual}
\end{figure}

For a quantitative evaluation (as shown in \figref{fig:closest_point_visual}), a standard Mean Squared Error (MSE) calculation is not ideal due to potential variations in the number of points between the generated and ground-truth trajectories, as well as the presence of preparation or finishing motions that are not critical to the core cutting action. 
To address these issues, the ground-truth motion was first downsampled to a fixed number of points to ensure a fair comparison. Then, for each point in the ground truth trajectory, we find the index \( j_i \) of the closest unpaired point in the generated trajectory and calculate the Euclidean distance \( d_i \) between them:
\begin{align}
    d_i = \| \mathbf{m}_i - \mathbf{n}_{j_i} \| \label{eq:eval_1}
\end{align}
where \( i = 1, 2, \dots, N \), with \( N \) being the number of points in the ground truth trajectory, \( \mathbf{m}_i \) is the \( i \)-th point in the ground truth trajectory, \( \mathbf{n}_{j_i} \) is the closest unpaired point in the generated trajectory to \( \mathbf{m}_i \), and \( j_i \) is the index of that point.

The overall evaluation metric \( D \) is computed as:
\begin{align}
    D = \frac{1}{N} \sum_{i=1}^{N} d_i \label{eq:eval_2}
\end{align}
This metric \( D \) provides a quantitative measure of the discrepancy between the ground truth trajectory and the generated trajectory, accounting for spatial accuracy and ensuring that each ground truth point is uniquely matched to a generated point. A lower value of \( D \) indicates a higher similarity between the generated motion and the ground truth, reflecting better performance of the method. This process enabled a consistent and accurate comparison between the generated and intended motions, focusing the comparison on the significant parts of the trajectory, excluding preparation or finishing motions. 

It is important to note that this metric evaluates geometric proximity and is therefore insensitive to the temporal sequence of the trajectory. An otherwise accurate set of points executed in an illogical or scrambled order could still yield a low error score, as path coherence is not assessed.

\paragraph{Primitive Dictionary Preparation}

We prepared two cutting primitives, selected from common motions in basic knife techniques \cite{weinstein2008mastering}, by having DMPs imitate predefined keypoints:
\begin{enumerate}
    \item Straight-downward cutting primitive [straight]: Involves a straight downward motion, suitable for soft objects requiring vertical cuts. The knife moves downward without significant horizontal motion.
    \item Sawing cutting primitive [sawing]: Incorporates a forward and backward sawing motion combined with downward force, ideal for harder objects needing more effort to cut through.
\end{enumerate}

To create primitives for the reference primitive dictionary, we designed basic task-related 3D trajectories to ensure smooth motion and then imitated these trajectories with DMPs to extract their parameters. 
Visualizations of these primitives are provided in Appendix D.

\subsubsection{Cake Icing Task}
\label{sim-exp-ice}

\paragraph{Task Description}

The objective of the cake icing task is to create icing designs on a cake according to the input prompt. 
The complexity of the designs depends on the intricacy of the desired outcome, impacting the generated keypoint pairs. 
Since the task is qualitative, we designed several cases shown in \tableref{table:icing-tasks} to assess the VLMs' capabilities in generating intricate icing designs.

\paragraph{Evaluation Method}

Given the qualitative nature of the cake icing task, evaluation was based on visual inspection and assessment of how well the generated designs aligned with the intended outcomes described in the input prompts. 
We examined whether the VLMs could produce complex keypoint pairs that resulted in creative and accurate icing designs, as specified in the task cases.

\paragraph{Primitive Dictionary Preparation}
For the cake icing task, the framework uses a single icing line primitive [line] designed to emulate standard decorating techniques \cite{garrett2004welldecorated}. 
The primitive's core motion involves a trajectory that first descends to the working surface. 
It then translates laterally to form a continuous line or decorative border.
A visualization of this motion is available in Appendix D.

This single-primitive approach streamlines the framework. 
It places greater emphasis on the VLM's ability to generate intricate keypoint designs. 
Its versatility is such that it can also produce icing dots (rosettes). 
This happens whenever the VLM specifies a keypoint pair with identical start and end coordinates.

\subsection{Implementation Details}

We implemented a pixel-based object detector to capture the object's global position (see Appendix A). 
A GPT-4o model \cite{openai_gpt4o_2024} initialized by the system prompts defined in \sref{primitive-selection} and \sref{pattern-generation} was utilized for the VLM components. 
Given that the same input is used in both components, we combined the system prompts for both components in the same VLM model, which outputted the results for both components.
The complete prompt used for these components is available on our project website (\href{https://keymps.github.io}{https://keymps.github.io}). 

The height of each object was directly measured, and a margin was added as needed to ensure safe spatial scaling parameters.
This approach allowed for accurate scaling of the cutting primitives while accommodating potential variations in object dimensions during execution.

\subsection{Experiment 1: Comparison with Direct VLMs-to-DMPs Approach}
\label{sim-exp2}

\begin{figure}[b]
\centering
\includegraphics[width=1.0\hsize]{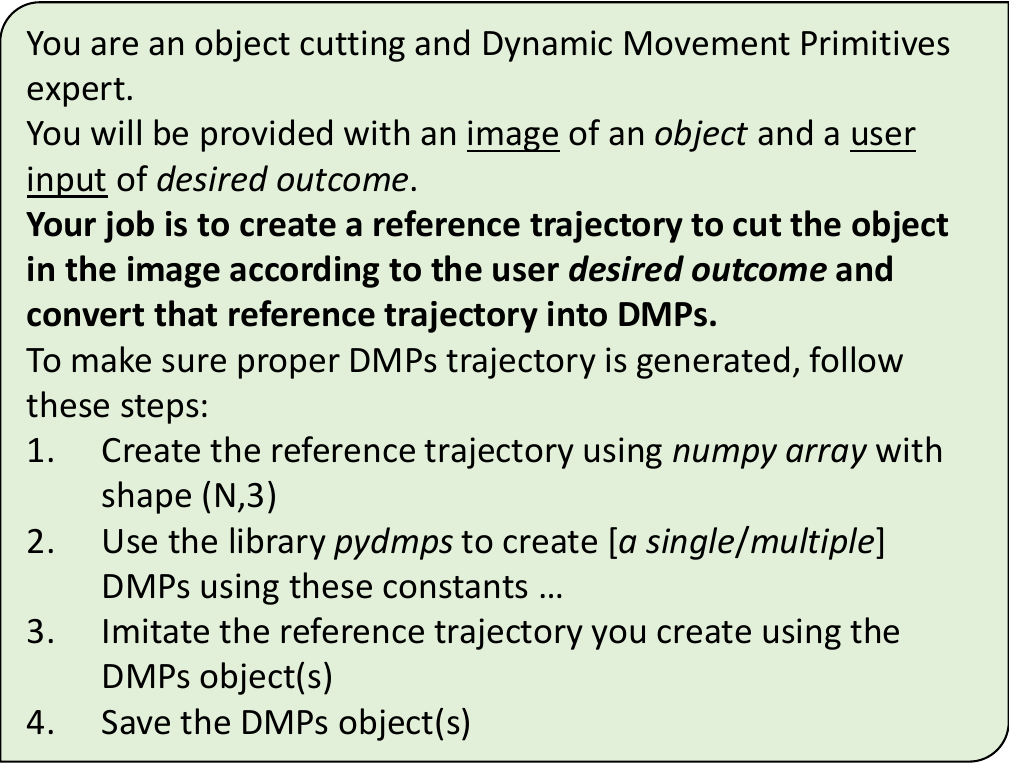}
\caption{
Direct VLM-to-DMPs initialization prompt.
}
\label{prompt:ex1}
\end{figure}

\begin{figure*}[t]
\centering
\includegraphics[width=1.0\hsize]{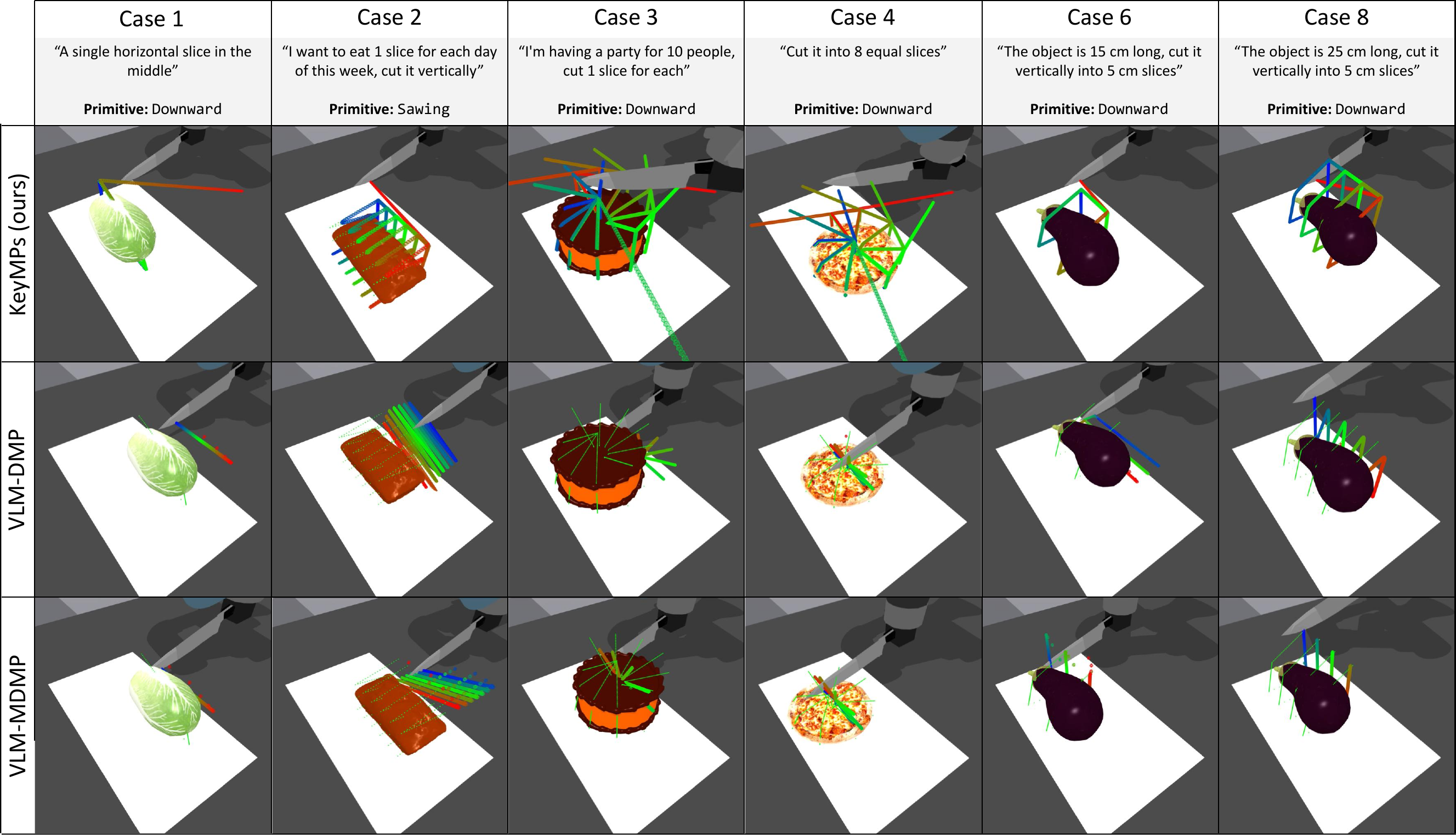}
\caption{Comparison of the cutting motions generated by KeyMPs and direct VLMs-to-DMPs methods. 
Time progression is depicted through a shift in color from red to green to blue. 
The thin green lines are ground-truth coordinates for evaluation purposes.}
\label{fig:ex_1_qual}
\end{figure*}

\begin{figure*}[t]
\centering
\includegraphics[width=1.0\hsize]{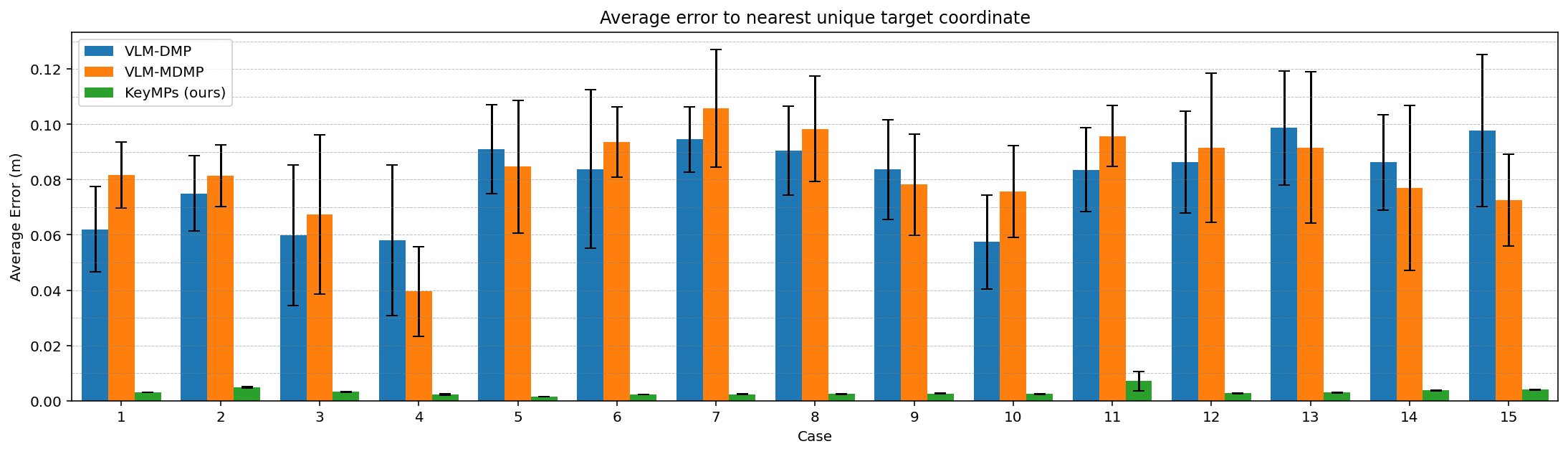}
\caption{Average error to nearest unique ground-truth coordinate results: comparison of KeyMPs against Direct VLMs-to-DMPs methods. 
KeyMPs significantly outperforms both VLM-DMP and VLM-MDMP, achieving notably lower error rates and reduced variance. 
These results underscore that simply increasing the number of DMPs without component decomposition does not overcome the inherent limitations of direct VLM-to-DMP approaches.}
\label{fig:ex_1_quan}
\end{figure*}

\begin{figure*}[t]
\centering
\includegraphics[width=1.0\hsize]{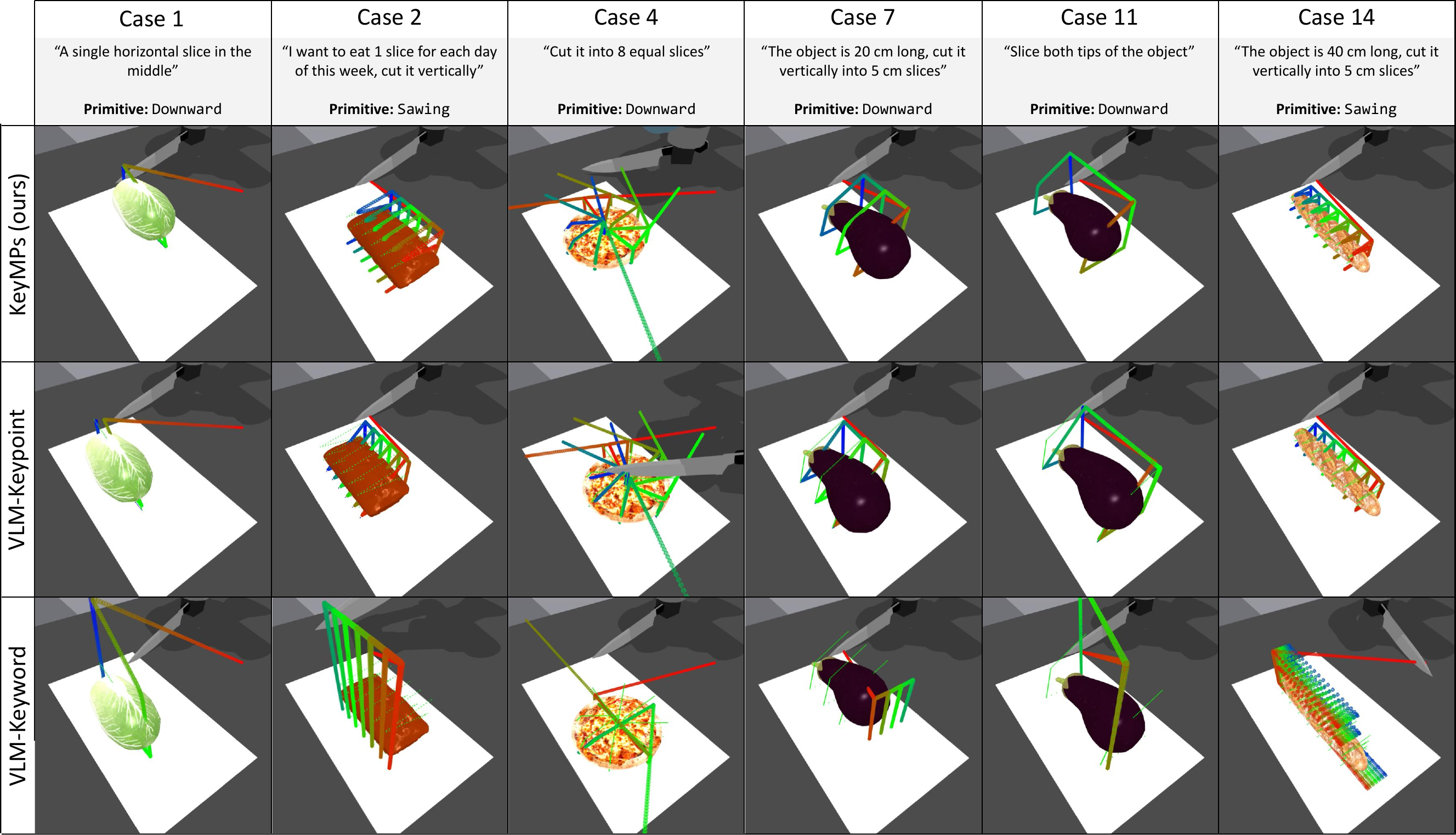}
\caption{Comparison of cutting motions generated by KeyMPs and ablation methods. 
Time progression is depicted through a shift in color starting from red to green to blue.
The thin green lines are ground-truth coordinates for evaluation purposes.}
\label{fig:ex_2_qual}
\end{figure*}

\begin{figure*}[h]
\centering
\includegraphics[width=1.0\hsize]{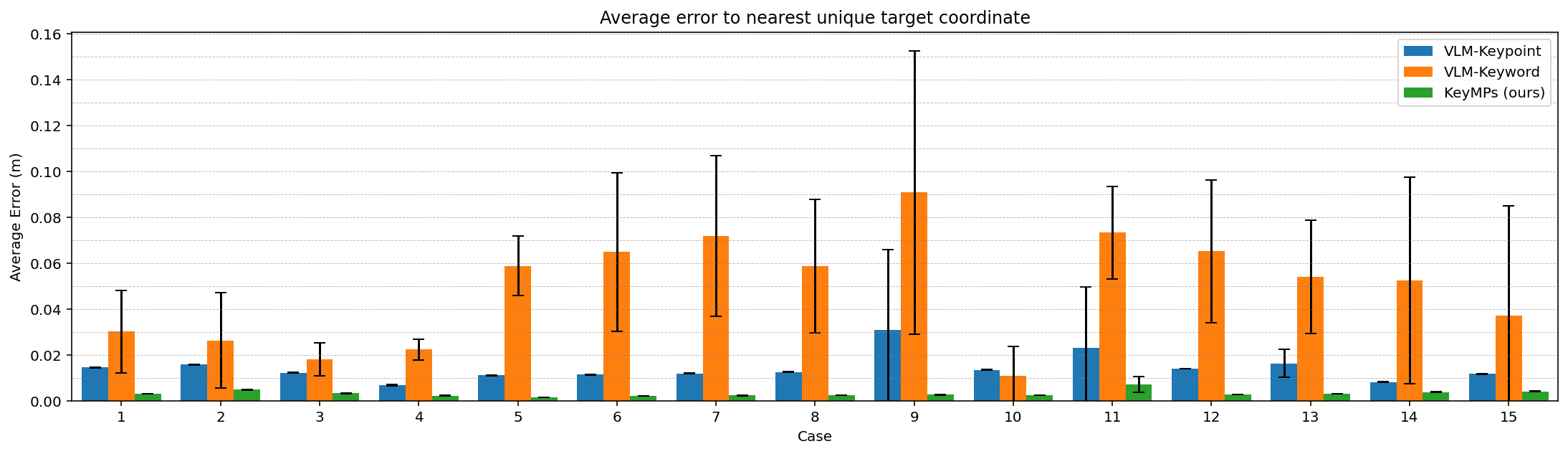}
\caption{Average error to nearest unique ground-truth coordinate result: comparison of KeyMPs against ablation methods. 
KeyMPs achieves the lowest error, followed by VLM-Keypoint showing how much it contributes to the overall framework.
In contrast, VLM-Keyword achieves the highest error, as it relies on the 3D keypoints directly generated by VLMs.}
\label{fig:ex_2_quan}
\end{figure*}

\begin{figure}[b]
\centering
\includegraphics[width=1.0\hsize]{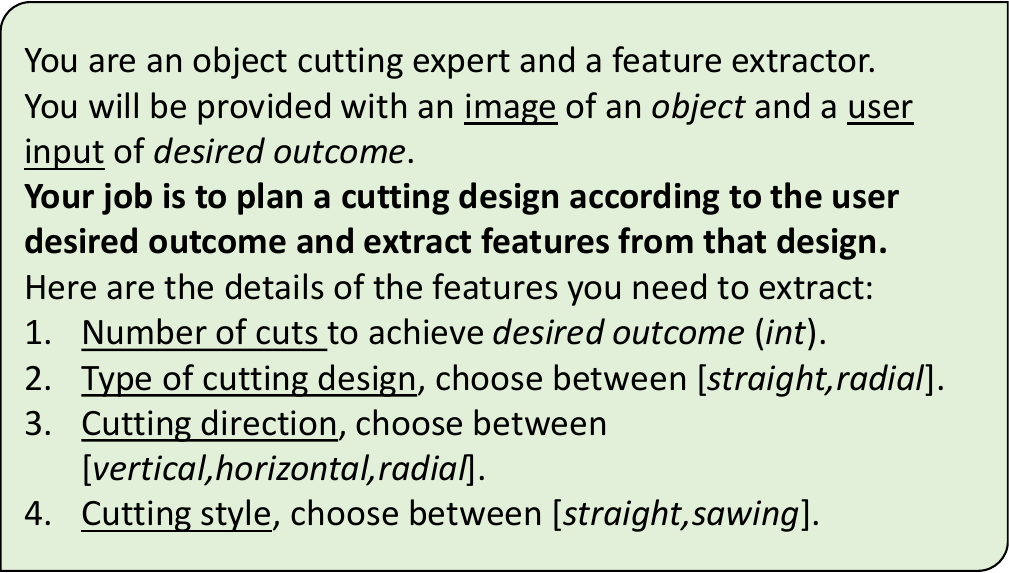}
\caption{
Task-feature-extractor initialization prompt.
}
\label{prompt:ex2}
\end{figure}

\subsubsection{Objective}

In this experiment, we evaluated the extent to which the structured component decomposition approach implemented in \textbf{KeyMPs} produces executable motions that align with the intent expressed in the multimodal input. 
Specifically, we compared \textbf{KeyMPs} against an unstructured, end-to-end VLM-driven approach that directly generates motion primitives. 
This comparison tested whether breaking the motion generation process into subtasks improves the capture of 3D geometric details and DMP parameters, resulting in motions that more faithfully reflect the intended outcome.
Each task in \tableref{table:cutting-tasks} was executed ten times and was quantitatively evaluated using the criteria in \sref{task-evaluation}.

\subsubsection{Comparison Methods}

\begin{table}[t]
\caption{\textbf{Features of the compared methods to address P1}}
\label{table:method-features}
\setlength{\tabcolsep}{3pt}
\begin{tabular}{|l|>{\centering\arraybackslash}p{1.36cm}|>{\centering\arraybackslash}p{1.36cm}|>{\centering\arraybackslash}p{1.36cm}|>{\centering\arraybackslash}p{1.36cm}|}
\hline
Method & Object Detector & Keypoints Generation & Primitive Selection & Multiple DMPs \\
\hline
KeyMPs (ours) & \checkmark & \checkmark & \checkmark & \checkmark \\
VLM-DMP & \checkmark & - & - & - \\
VLM-MDMP & \checkmark & - & - & \checkmark \\
\hline
\end{tabular}
\end{table}

We compared our framework (\textbf{KeyMPs}) with two ablation methods that rely on direct VLM-to-DMP generation. 
The first method, \textbf{VLM-DMP}, employed single DMP generation without additional component decomposition, while the second, \textbf{VLM-MDMP}, used multiple DMP generation under similar unstructured conditions. 
Both methods were designed to process the same input as our proposed framework and produce Python code that, when executed, generates the 3D motion for the robot to execute.

A summary of the key features of each approach is provided in \tableref{table:method-features}. 
The system prompt used to generate DMP parameters directly from VLMs is shown in \figref{prompt:ex1}, and we utilized the Python library pydmps \cite{pydmps} to facilitate DMP motion reconstruction.

\subsubsection{Results}

Without a reference primitive, the direct VLM-to-DMP methods struggled to generate proper 3D motions, as shown in \figref{fig:ex_1_qual}. 
Both \textbf{VLM-DMP} and \textbf{VLM-MDMP} tended to produce primarily 2D-like trajectories, often with incorrect orientations. 
\textbf{VLM-DMP} was particularly limited by its constraint of generating a single connected motion, while \textbf{VLM-MDMP}, despite producing multiple sets of DMP parameters, still suffered from the inherent inability of VLMs to capture full 3D details. 
In contrast, \textbf{KeyMPs} circumvented these issues by asking VLMs to generate only 2D keypoint pairs and then using reference primitives to create detailed 3D motions.

The quantitative results support these observations. 
As depicted in \figref{fig:ex_1_quan},\textbf{KeyMPs} significantly outperformed both \textbf{VLM-DMP} and \textbf{VLM-MDMP}, achieving notably lower error rates and reduced variance. 
The direct VLM-to-DMP approaches exhibited high variability between inferences, with \textbf{VLM-MDMP} showing only marginal improvements over \textbf{VLM-DMP}, confirming that simply increasing the number of DMPs does not deal with the underlying limitations.

Overall, these findings demonstrate that \textbf{KeyMPs} effectively addresses (P1) by generating executable motions that align with the intended outcome expressed in the multimodal input. 
By structurally decomposing the motion generation process—using VLMs for high-level task understanding and 2D keypoint generation while relying on reference primitive for detailed 3D reconstruction—\textbf{KeyMPs} delivers precise and consistent motions that meet the desired outcomes.

\begin{figure*}[t]
\centering
\includegraphics[width=1.0\hsize]{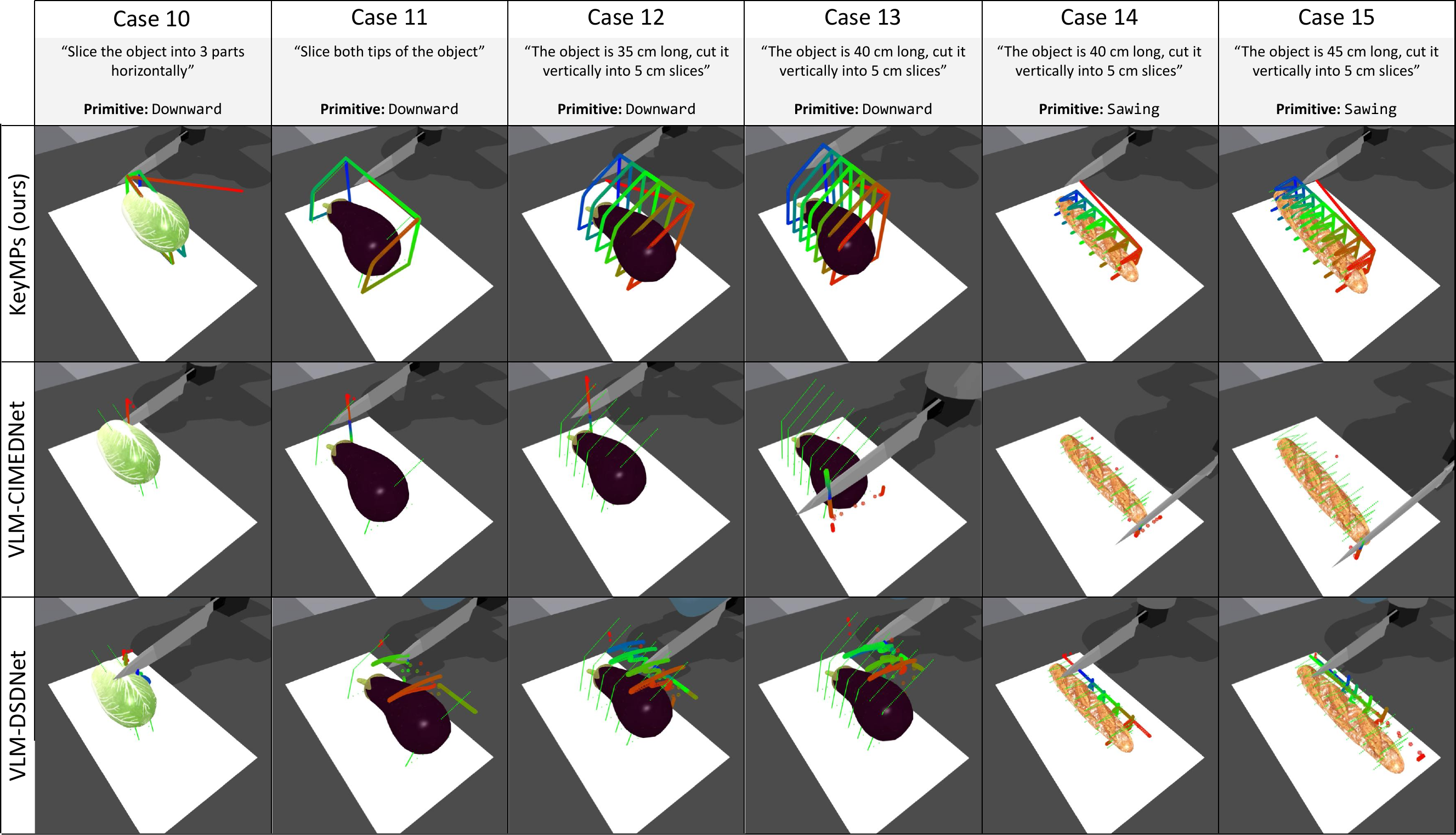}
\caption{Comparison of cutting motions generated by KeyMPs and learning based methods on unseen cases. Time progression is depicted through a shift in color starting from red to green to blue.
The thin green lines are ground-truth coordinates for evaluation purposes.}
\label{fig:ex_3_qual}
\end{figure*}

\begin{figure*}[h]
\centering
\includegraphics[width=1.0\hsize]{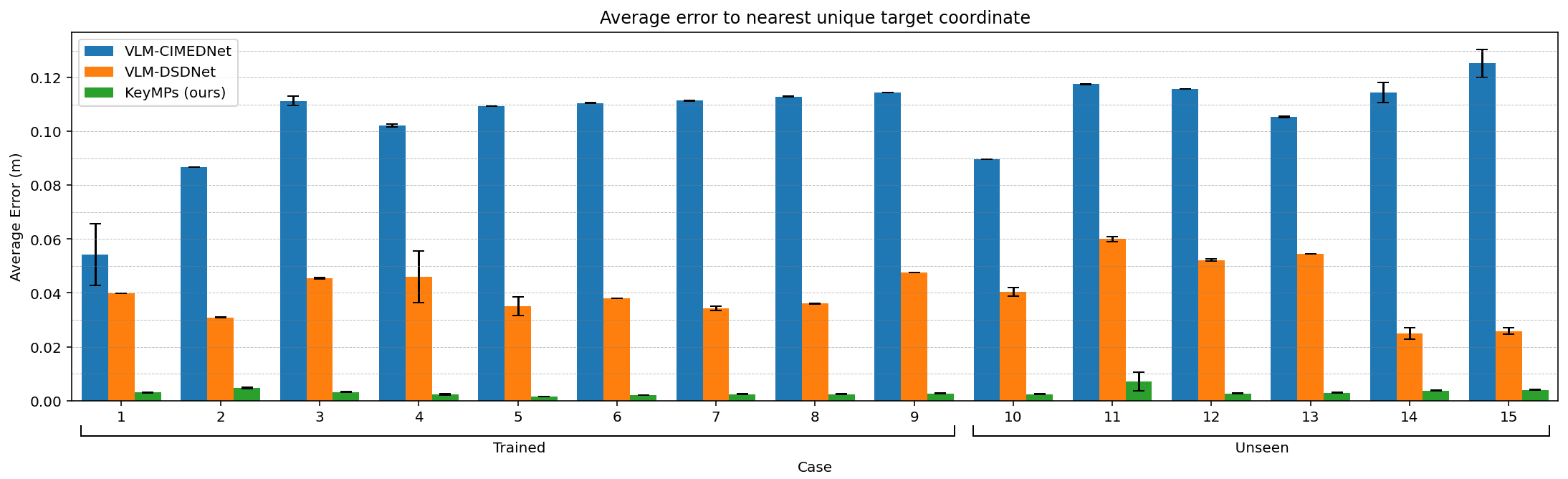}
\caption{Average error to nearest unique ground-truth coordinate result: comparison of KeyMPs against deep-learning approaches. 
KeyMPs exhibits significantly lower error rates and more consistent performance than either VLM-CIMEDNet or VLM-DSDNet. 
VLM-DSDNet tends to overfit to the training data while VLM-CIMEDNet fails to generate correct motions for unseen tasks.}
\label{fig:ex_3_quan}
\end{figure*}

\subsection{Experiment 2: Ablation of VLM-based Components}
\label{sim-exp3}

\subsubsection{Objective}

To address (P2), we evaluated the individual impacts of the two VLM-based components in our framework, i.e. keypoint pairs generation and keyword labeled primitive selection. 
This ablation study isolated each component's effectiveness in producing executable motions aligned with the intended outcome. 
Each task in \tableref{table:cutting-tasks} was executed ten times and assessed using the criteria in \sref{task-evaluation}.

\subsubsection{Comparison Methods}

\begin{table}[t]
\caption{\textbf{Features of the compared methods to address P2}}
\label{table:method-features-ablation}
\setlength{\tabcolsep}{3pt}
\begin{tabular}{|l|>{\centering\arraybackslash}p{1.36cm}|>{\centering\arraybackslash}p{1.36cm}|>{\centering\arraybackslash}p{1.36cm}|>{\centering\arraybackslash}p{1.36cm}|}
\hline
Method & Object Detector & Keypoints Generation & Primitive Selection & Multiple DMPs \\
\hline
KeyMPs (ours) & \checkmark & \checkmark & \checkmark & \checkmark \\
VLM-Keypoint & \checkmark & \checkmark & - & \checkmark \\
VLM-Keyword & \checkmark & - & \checkmark & \checkmark \\
\hline
\end{tabular}
\end{table}

We assessed each component's contribution by disabling the other. 
In \textbf{VLM-Keypoint}, we activated only keypoint pairs generation, setting all basis-function weights \( w \) in the DMPs to zero, effectively disabling the forcing function.
The final motion was then generated by sequencing the resulting DMPs based on the keypoint pairs.
In \textbf{VLM-Keyword}, we bypassed keypoint generation, instead directly prompting the VLMs to generate 3D keypoints without using a more structured component decomposition while retaining the keyword labeled primitive selection to select the primitive to be scaled by the generated 3D keypoints, and generated the final motion by sequencing DMPs. 

The features of each method are summarized in \tableref{table:method-features-ablation}.
This table highlights the availability of keypoints generation and primitive selection on the methods used in this experiment, enabling a direct comparison of their individual contributions to motion generation in the ablation study.

\subsubsection{Results}

\figref{fig:ex_2_qual} shows qualitative results that expose the limitations of the ablated methods.
\textbf{VLM-Keypoint}, lacking $w$, produced straight-line motions between keypoints, resulting in oversimplified, less detailed trajectories. 
\textbf{VLM-Keyword} mirrored the issues from \sref{sim-exp2}, often generating flat 2D keypoints or misoriented 3D keypoints. 
Even in cases where it accurately generated the cutting keypoints with the correct orientation, it still struggled to predict the height of the object and failed to capture essential 3D geometric details.

\figref{fig:ex_2_quan} presents quantitative results that reinforce these findings.
\textbf{VLM-Keypoint} achieved lower error, nearly matching \textbf{KeyMPs} and highlighting the critical role of keypoint pairs generation in enhancing motion generation. 
\textbf{VLM-Keyword} showed significantly higher error and variance, reflecting the unreliability of directly generated 3D keypoints. 
In regard to (P2), these results demonstrate that keypoint pairs generation contributed significantly to motion accuracy, while keyword labeled primitive selection ensured consistency, and that, together in the form of \textbf{KeyMPs}, they achieved the lowest error and minimal variance across all scenarios for precise motion generation.

\subsection{Experiment 3: Comparison with Deep-Learning Approach}
\label{sim-exp1}

\subsubsection{Objective}

To address (P3), we determined whether \textbf{KeyMPs} achieves better generalization to unseen cutting tasks than deep-learning methods when training data is limited. 
The deep-learning baselines were trained on a restricted dataset of 90 demonstrations (10 per task for 9 tasks in \tableref{table:cutting-tasks}), simulating scenarios where collecting large-scale robotic data is impractical. 
In contrast, \textbf{KeyMPs} doesn't require extensive task-specific training data, relying instead on structured component decomposition and vision-language priors. 
Each task in \tableref{table:cutting-tasks} was executed ten times, and the quantitative evaluation (\sref{task-evaluation}) focused on generalization performance, specifically success rates on novel objects and cutting patterns.

\subsubsection{Comparison Methods}

We compared our framework with modified versions of CIMEDNet \cite{pahic2020training} and DSDNet \cite{anarossi2023deep}, both of which are convolutional autoencoder-based deep learning models designed to predict DMP parameters from visual inputs. 
They were trained on Cases 1 through 9 of \tableref{table:cutting-tasks}. 
CIMEDNet predicts a single set of DMP parameters that are typically used to generate a DMP with a large number of basis functions for motion imitation, i.e., aiming to capture detailed trajectories in a single motion primitive. 
In contrast, DSDNet predicts multiple sets of DMP parameters that can be sequenced to represent an overall complex motion in a way that reduces the need for training deep learning model extensive amounts of  data. 

We modified both methods by replacing their encoder layers, which traditionally perform dimensionality reduction, with VLM prompts (see \figref{prompt:ex2}) designed to extract relevant task information directly from multimodal inputs. 
Both methods were further enhanced by integrating an object detector to minimize positional variability and facilitate easier learning of DMP parameters.
They are referred to as \textbf{VLM-CIMEDNet} and \textbf{VLM-DSDNet}.

\subsubsection{Results}

As illustrated in \figref{fig:ex_3_qual}, the deep-learning-based method \textbf{VLM-CIMEDNet} completely failed to generate the correct cutting motion in the unseen tasks due to the vast feature space required for each motion. 
Accordingly, our discussion will focus on \textbf{VLM-DSDNet}, which, while it was able to reproduce cutting motions to some extent, it still exhibited significant deficiencies. 
Specifically, we identified three generalization groups based on the qualitative outcomes: Group 1 (Cases 1, 5, 6, and unseen Case 10) where, despite receiving appropriate high-level input, \textbf{VLM-DSDNet} erroneously repeated vertical cuts instead of adapting to a mixed cutting pattern; 
Group 2 (Case 11) where the model should produce two vertical cuts near the object’s tips but instead distributed them evenly across the object;
Group 3 (Cases 2, 5–9 and unseen Cases 12–15) where the model is expected to extrapolate the correct number of cuts and adjust their spacing according to the object’s dimensions, but overfit to the training data (notably in Cases 7 and 8) and failed to adapt its output for longer objects, resulting in repetitive patterns that did not meet the varying requirements.

The quantitative analysis, shown in \figref{fig:ex_3_quan}, further supports these observations. 
The error metrics for \textbf{VLM-CIMEDNet} were consistently high, reflecting its inability to effectively learn the DMP parameters. 
For \textbf{VLM-DSDNet}, although the variance was high across all unseen tasks, with only marginally lower errors on tasks resembling the training set (e.g., Cases 7 and 8), the errors increased significantly for tasks that diverged from these overfitted patterns. 
Notably, \textbf{KeyMPs} achieved significantly lower error values across both trained and unseen tasks, highlighting its robust generalization without the limitations observed in the deep-learning approaches.

Overall, these findings demonstrate that while \textbf{VLM-DSDNet} was partially successful in reproducing the learned cutting patterns, it struggled to generalize accurately to unseen tasks, overfitting to the training data. 
In addition, \textbf{VLM-CIMEDNet} completely failed to generate correct motions for unseen tasks due to its inability to handle the large feature space required by each motion. 
In contrast, \textbf{KeyMPs} leveraged structured component decomposition and VLMs' knowledge and achieved lower error rates and better generalization, thereby effectively addressing (P3).

\subsection{Experiment 4: Evaluating Framework Flexibility for Broader Applications}
\label{sim-exp4}

\subsubsection{Objective}

To answer (P4), we test the flexibility of our framework by applying it to a new application: cake icing. 
This task was chosen to assess whether the core principles of KeyMPs' VLM driven keypoint generation combined with a single motion primitive can generalize to a broader variety of creative and intricate tasks, with scenarios listed in Table 4.

\subsubsection{Results}

\begin{table}[t]
\caption{\textbf{List of prepared cake icing tasks in the real environment}}
\label{table:icing-tasks}
\setlength{\tabcolsep}{3pt}
\begin{tabular}{|l|c|>{\centering\arraybackslash}p{6cm}|}
\hline
Case & Object & Input Prompt \\
\hline
1 & Round cake & Put 15 icings around the cake, provide some space from the cake edge and between each icing. \\
2 & Round cake & Make 2 layers of icing around the cake, provide some space from the edges and between layers. \\
3 & Round cake & Combine 2 types of designs, icing dots around the cake and a spiral icing line design. \\
4 & Round cake & I'm having my 27th birthday, put the digits on the cake using seven-segment characters as icing lines. \\
5 & Square cake & Draw a 5-pointed star using lines in the middle of the cake. \\
6 & Round cake & Create a simple interesting design. \\
7 & Square cake & Create a simple interesting design. \\
8 & Square cake & I want to put 1 dot of icing on the middle of each of the 10 slices. \\
\hline
\end{tabular}
\end{table}

\begin{figure}[t]
\centering
\includegraphics[width=1\linewidth]{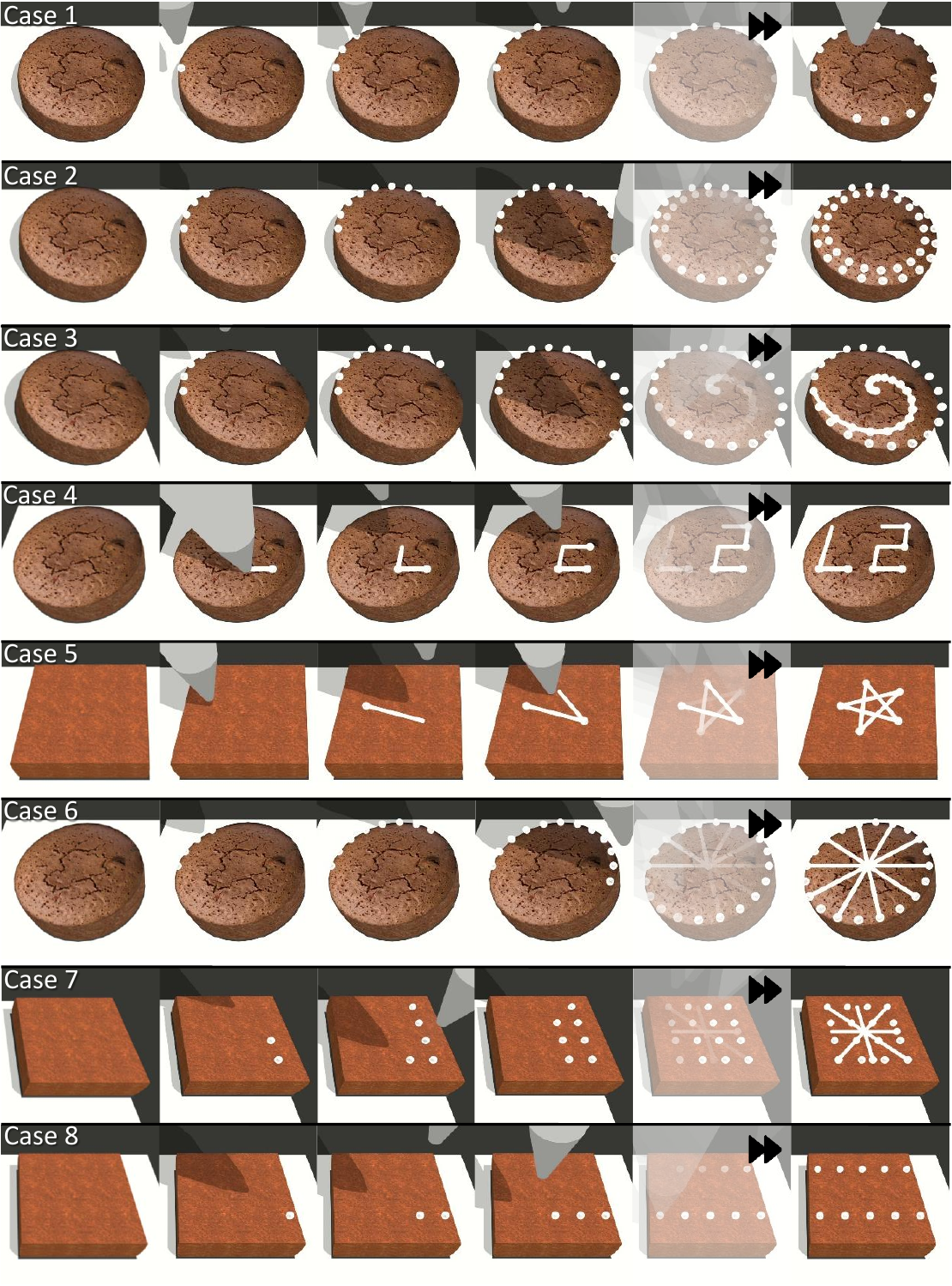}
\caption{
Results of the cake icing experiments, demonstrating the generalizability of the KeyMPs framework. 
Right-most column shows the final outcome for the corresponding task case in \tableref{table:icing-tasks}. 
}
\label{fig:icing-result}
\end{figure}

As shown in \figref{fig:icing-result}, the framework demonstrated significant flexibility across all scenarios. 
The system generated simple geometric patterns (Cases 1, 2), combined different design elements (Case 3), and produced motions from abstract concepts like numbers and shapes (Cases 4, 5).

Furthermore, the framework handled ambiguous prompts by generating creative yet context-aware designs (Cases 6, 7). 
For example, when tasked with creating a "simple interesting design," it produced a circular pattern for the round cake and a square-based pattern for the square cake. 
The framework also solved multi-step reasoning problems, such as dividing a cake before placing icing (Case 8).

Collectively, these outcomes show that the KeyMPs architecture has the potential for broader application. 
By simply providing a new primitive and leveraging the VLMs' contextual understanding, the framework successfully generates motions for this distinct application. 
This directly answers (P4) by demonstrating the framework's capacity for generalization to new tasks.

\section{REAL ROBOT EXPERIMENT}

Building on the experimental setup detailed in \sref{sim-exp-cut}, we further validated the effectiveness of the \textbf{KeyMPs} framework in a real-world setting. 
This experiment assessed if the generated motions align with the multimodal input intent and evaluated the framework’s practical performance.

\subsection{Research Questions}

We addressed the following key question:

\begin{itemize}
    \item[(P5)] Can \textbf{KeyMPs} generate effective and practical executable motions from images in a real-world setting? \sref{real-exp3}
\end{itemize}

\subsection{Experimental Setting}

\subsubsection{Task Description}

\begin{figure}[t]
\centering
\includegraphics[width=1.0\hsize] {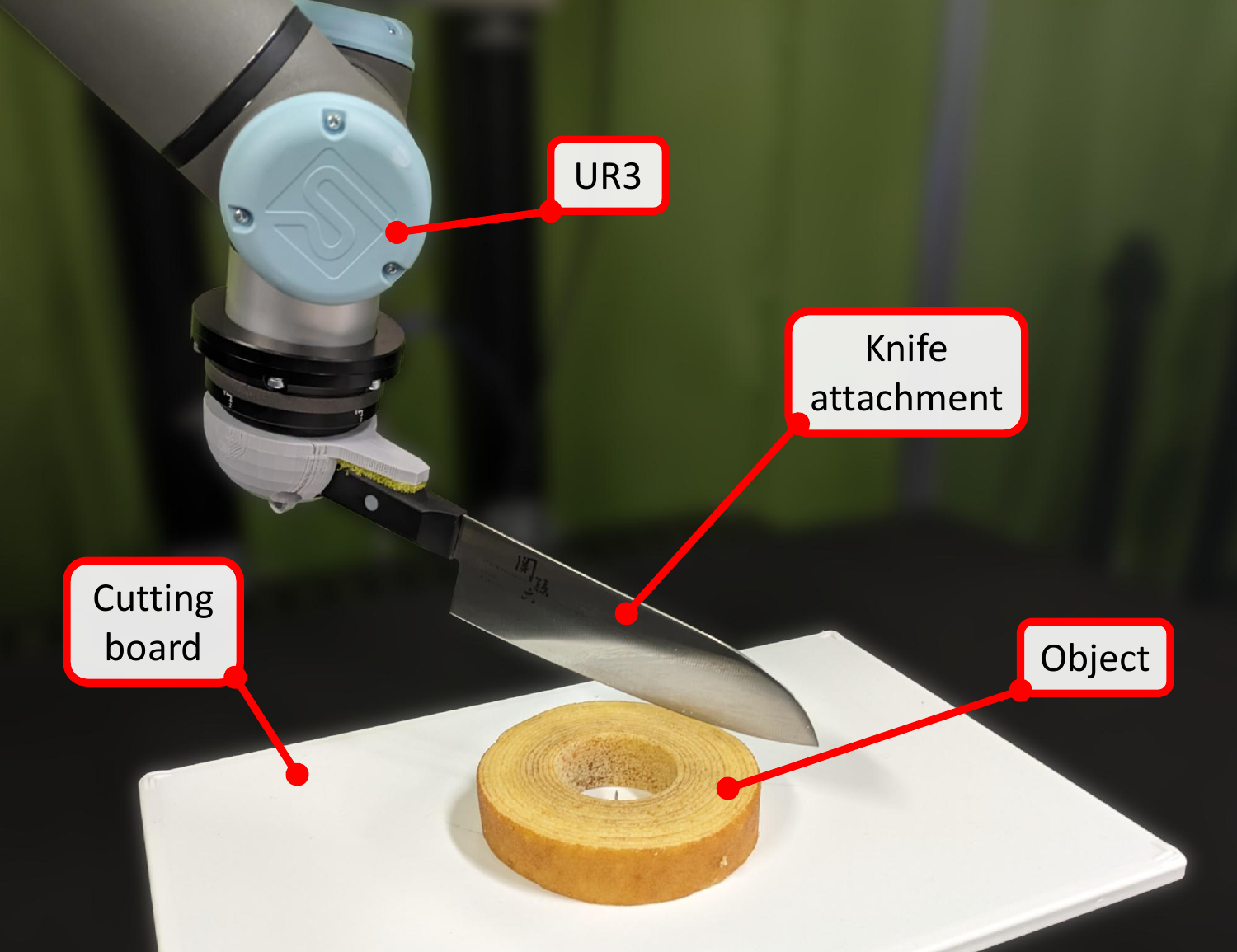}
\caption{
Object cutting environment in real-world setting.
}
\label{fig:real_env}
\end{figure}

The goal, similar to the one of the simulation, was to cut objects (listed in \tableref{table:cutting-real-tasks}) on a $\SI{0.24}{\meter} \times \SI{0.38}{\meter}$ cutting board based on user input. 
The experimental robot environment is shown in \figref{fig:real_env}.

We used a 6-DOF Universal Robot 3 (UR3) equipped with a knife attachment and an RGB webcam (Logitech Webcam C615 HD) for image observation of the cutting-board area. 
The webcam was mounted approximately $\SI{1.15}{\meter}$ directly above the cutting board to capture a top-down view. 
Nails were installed on the cutting board to stabilize the objects, and a wooden spatula was employed during motion execution to prevent any object displacement.

We designed ten task cases (see \tableref{table:cutting-real-tasks}) to assess the performance of our framework in a real-world setting. 
Each task was executed, and the outcomes were qualitatively evaluated on the basis of successful completion of the cutting tasks as per the user's input.

\begin{table}[b]
\caption{\textbf{List of prepared cutting tasks in the real environment}}
\label{table:cutting-real-tasks}
\setlength{\tabcolsep}{3pt}
\begin{tabular}{|l|c|>{\centering\arraybackslash}p{6cm}|}
\hline
Case & Object & Input Prompt \\
\hline
1 & Chiffon cake & I have 3 guests, cut a few thin slices of the chiffon cake for them. \\
2 & Chiffon cake & I want to eat this chiffon cake for each day this week. \\
3 & Cucumber & I want to make tsukemono. \\
4 & Eggplant & I want to make wide chips out of this. \\
5 & Baumkuchen & Split it into 4. \\
6 & Meat loaf & This is a 490g block of meat (length 21cm, width 8cm), the nutrition facts mentioned that every 100g there's 150kcal. Cut me the several number of slices in a certain length (below 3 cm) just enough if I want to go for a 3km walk after this. \\
7 & Meat loaf & This is a block of meat (length 16cm, width 8cm). Cut me 4 2cm slices. \\
8 & Potato & Prepare it for fondant potato, this potato is quite small. \\
9 & Sliced Potato & Cut it into french fries. \\
10 & 2 bananas & For banana pancake. \\
\hline
\end{tabular}
\end{table}

\subsubsection{Primitive Dictionary Preparation}

For the real robot experiment, we prepared two types of cutting primitive by having DMPs imitate predefined keypoints:
\begin{enumerate}
    \item \textbf{Downward cutting primitive [downward]}: Used for soft objects.
    \item \textbf{Forward cutting primitive [forward]}: Suitable for harder objects.
\end{enumerate}
Initially, a sawing primitive used in §V was utilized. 
However, due to displacement resulting from this motion, the primitive was simplified to a forward cutting primitive to mitigate displacement.
Visualizations of primitives used in the real experiments are presented in Appendix D.

\subsection{Experiment 5: Feasibility in a Real-World Setting}
\label{real-exp3}

\subsubsection{Objective}

The goal of this experiment was to assess whether \textbf{KeyMPs} can generate effective, executable motions in a real-world setting by using more realistic vision and language input. 
Specifically, we aimed to determine if our framework, by processing actual visual input from an RGB webcam alongside more complex language instructions, would produce DMP-based motions that faithfully reflected the intended outcome. 
This evaluation addressed (P5) by testing the feasibility and consistent performance of \textbf{KeyMPs} under practical conditions, where sensor noise might be present.

\begin{figure*}
\centering
\includegraphics[width=0.89\hsize]{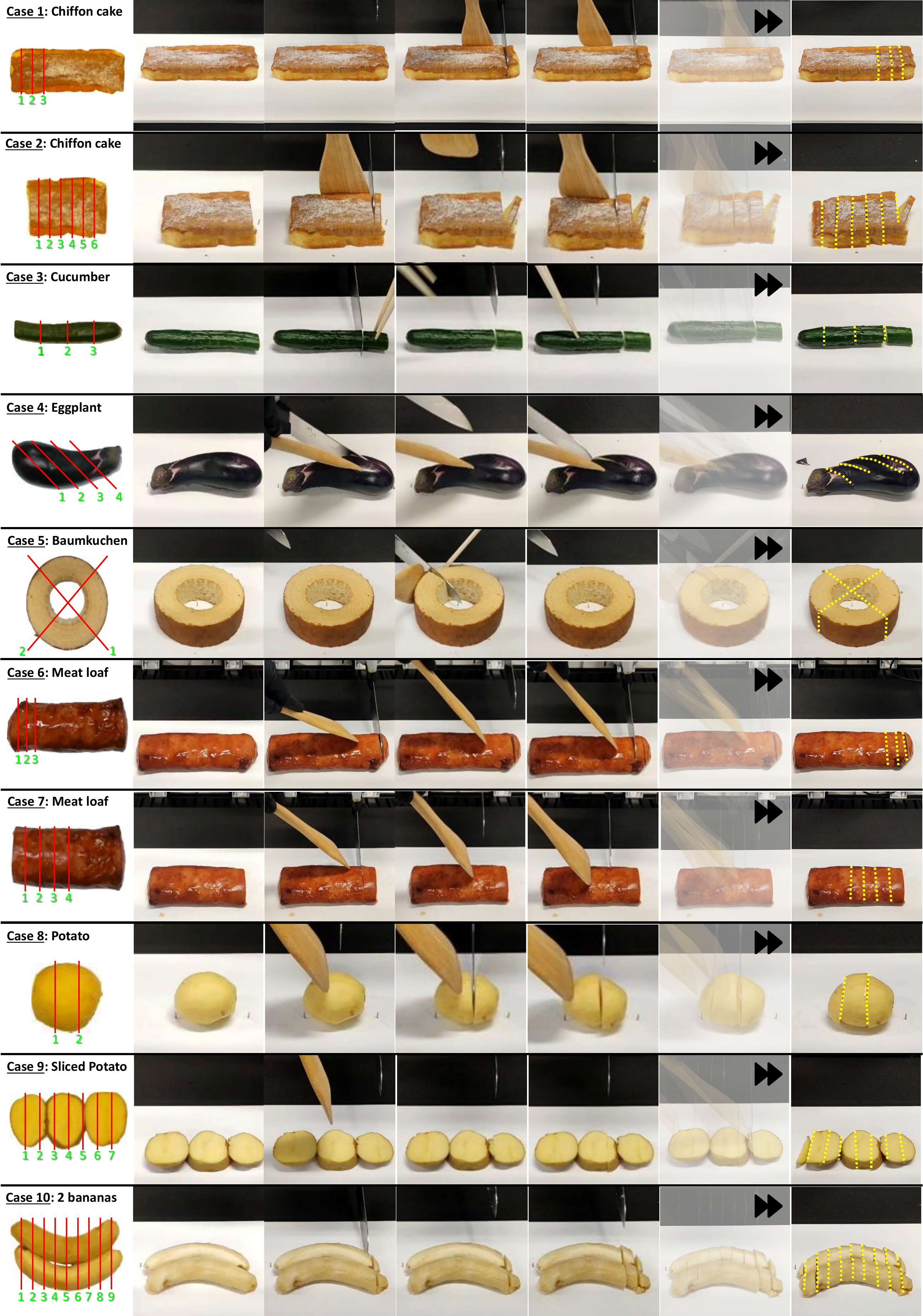}
\caption{
Results of real robot experiment for all task cases in \tableref{table:cutting-real-tasks}. 
Red lines in the first column represent the visualization of the VLMs generated keypoint pairs with green numbers for the order of cuts. 
In the last column, the yellow-dotted lines represent how the objects are cut.
} 
\label{fig:cutting-result}
\end{figure*}

\subsubsection{Results}

Time-lapse recordings of the real robot experiments, shown in \figref{fig:cutting-result}, clearly demonstrated that \textbf{KeyMPs} generated practical executable motions from real images in a real-world setting. 
The evolving motion sequences illustrate how the system successfully translates multimodal inputs—combining live visual data with language instructions—into detailed cutting actions. 
Video demonstrations available on our project website (\href{https://keymps.github.io}{https://keymps.github.io}) further confirm that the generated motions reliably align with the intended outcomes under realistic conditions.

In Case 6, the VLMs exhibited advanced reasoning by determining the optimal number of meat slices based on caloric requirements for a 3\,km walk, showcasing their capability of handling complex tasks. 
Cases 9 and 10 further highlight the system’s versatility in managing multiple objects simultaneously, such as cutting potatoes into french fries and processing two bananas for a pancake recipe.

Overall, these results demonstrate that \textbf{KeyMPs} effectively addresses (P5) by generating executable motions that not only match the intent expressed in the multimodal input but also adapt to real-world sensor noises. 
By leveraging VLMs for high-level task understanding and 2D keypoint generation while relying on reference primitive for detailed 3D reconstruction, \textbf{KeyMPs} produces consistent and reliable motions. 
This confirms the framework’s potential for practical deployment in robotic applications that require both deep reasoning and consistent performance in real-world settings.

\section{DISCUSSION}

The results of the evaluation confirm that our \textbf{KeyMPs} framework generates complex executable motions in one shot that closely align with the intent expressed in the multimodal input, while also achieving consistent and data-efficient DMP-based motion generation in occlusion-rich tasks. 
Moreover, an additional experiment detailed in Appendix F demonstrates that integrating multimodal input significantly outperforms using language input alone, even when supplemented with additional context. 
By effectively handling learned and spatial scaling parameters separately and leveraging VLMs' conceptual understanding and high-level reasoning capabilities, the framework requires only a single demonstration for each primitive.

Despite these advantages, the framework has several limitations:
\begin{enumerate*}
    \item Dependence on a Predefined Primitive Dictionary: KeyMPs relies on a predefined primitive dictionary and expert-crafted primitives, which limits its autonomy and scalability. We did not use primitives from real-world demonstrations, each primitive had to be handcrafted by an expert, requiring specialized expertise and manual effort.
    \item Limited Handling of Multi-Object Interactions: The current approach is designed for tasks involving a single cluster of objects and does not accommodate interactions between multiple objects with varying positions.
    \item Absence of Temporal Scaling: The framework focuses on the geometric generation of robotic motion and does not incorporate the temporal scaling parameter \( \tau \), which is crucial for adjusting the motion's execution speed.
    \item Reliance on Domain-Specific Post-Processing: Post-processing of the VLM-generated 2D keypoint pairs into usable spatial scaling parameters depends on domain-specific knowledge and manual oversight, which reduces adaptability to new tasks without expert input.
\end{enumerate*}

Future work can address these limitations by:
\begin{enumerate*}
    \item Automating Primitive Generation: Reducing reliance on human demonstrations is crucial for full autonomy. 
    Generating primitives directly through VLMs or other automated processes could simplify creation, reduce human involvement, and enhance scalability across tasks.
    \item Extending \textbf{KeyMPs} to Multi-Object Tasks: Expanding the framework to handle interactions between multiple objects by utilizing VLMs' ability to reason about complex scenes and relationships would increase its applicability to diverse and dynamic environments.
    \item Leveraging VLMs for Temporal Scaling: Utilizing VLMs' embedded knowledge about tasks and object properties could enable automatic adjustment of the motion's temporal scaling \( \tau \), broadening the framework's applicability to tasks requiring precise timing.
    \item Enhancing Post-Processing of VLMs output: Improving post-processing to be more generalizable and less dependent on domain-specific knowledge. 
    Leveraging VLMs' embedded knowledge further could automate and simplify post-processing, thereby reducing reliance on human expertise and broadening the framework's applicability to more complex tasks.
\end{enumerate*}

\section{CONCLUSION}
This paper introduced \textbf{KeyMPs}, a novel framework that enhances robotic motion generation through the utilization of VLMs and sequencing DMPs.
By leveraging VLMs' high-level reasoning for selecting reference DMPs and their spatial awareness for keypoint pairs generation, KeyMPs effectively bridges different high-level language instructions with motion generation through sequencing DMPs.
The structured decomposition for sequencing multiple DMPs enables 
one-shot vision-language guided motion generation that is adaptable and generalizable in occlusion-rich tasks while reducing the need for extensive human demonstrations. 
Validated through both simulation and real-world experiments, KeyMPs consistently produced motions that accurately aligned with the intent expressed in the multimodal input, achieving a high degree of consistency and data efficiency in multiple tasks.

\section*{APPENDIX}
\section*{A. Pixel-Based Object Detection}

To detect the object within the environment observation image, we implement an object detection method based on pixel intensity thresholds. 
The general flow of the method is as follows:

\begin{enumerate}
    \item \textbf{Pre-Processing:}
    \begin{itemize}
        \item \emph{Image Acquisition and Conversion}: Load the environment observation image and convert it to grayscale to simplify processing.
        \item \emph{Noise Reduction}: Apply a Gaussian filter to the grayscale image to reduce noise and smooth out intensity variations.
    \end{itemize}
    \item \textbf{Background Estimation and Thresholding:}
    \begin{itemize}
        \item \emph{Background Intensity Estimation}: Identify the most common pixel intensity value in the smoothed grayscale image, which represents the background intensity.
        \item \emph{Object Mask Creation}: Define a threshold based on the background intensity. Pixels with intensity values differing from the background by more than this threshold are considered part of the object, resulting in a binary object mask.
    \end{itemize}
    \item \textbf{Bounding Box Extraction:}
    \begin{itemize}
        \item \emph{Contour Detection}: Detect contours in the object mask using contour-finding algorithms.
        \item \emph{Largest Contour Selection}: Select the largest contour, assuming it corresponds to the object of interest.
        \item \emph{Bounding Box Calculation}: Compute a bounding box around the largest contour, providing the object's position and size in pixel-space coordinates.
    \end{itemize}
    \item \textbf{Output:} Return the coordinates of the bounding box, effectively capturing the object's position in the pixel-space for further processing.
\end{enumerate}

This method efficiently extracts the object's bounding box on the basis of significant pixel intensity differences, facilitating subsequent steps in the motion generation pipeline.

\section*{B. Transforming 2D Keypoint Pairs from Local to Global Coordinates}

After obtaining the keypoint pairs in the image's local coordinates, we need to transform them into global coordinates that align with the environment's coordinate system. 
This transformation involves translating and scaling the keypoints based on the object's bounding box obtained from the previous step.

Let us define the following terms describing the process of translating and scaling keypoint pairs:
\begin{itemize}
    \item \( \mathbf{p}_{\text{local}} = (x_{\text{local}}, y_{\text{local}}) \) is a keypoint in the local coordinate system.
    \item \( \mathbf{p}_{\text{global}} = (x_{\text{global}}, y_{\text{global}}) \) is the corresponding keypoint in the global coordinate system.
    \item \( \mathbf{b}_{\text{offset}} = (x_{\text{offset}}, y_{\text{offset}}) \) is the offset of the object's bounding box in the image.
    \item \( \mathbf{g}_{\text{shift}} = (x_{\text{shift}}, y_{\text{shift}}) \) is the shift from the environment's origin to the global coordinate system.
    \item \( \mathbf{s}_{\text{img}} = (w_{\text{img}}, h_{\text{img}}) \) is the size of the image.
    \item \( \mathbf{s}_{\text{env}} = (w_{\text{env}}, h_{\text{env}}) \) is the size of the environment limits.
\end{itemize}

The transformation from local to global coordinates involves the following steps:

\begin{enumerate}
    \item \textbf{Translation:} Adjust the keypoint by both the bounding box offset and the global coordinate shift:
    \begin{equation}
        \mathbf{p}' = \mathbf{p}_{\text{local}} + \mathbf{b}_{\text{offset}} + \mathbf{g}_{\text{shift}},
    \end{equation}
    where \(\mathbf{g}_{\text{shift}}\) accounts for any displacement between the environment's origin and the actual global coordinate system.

    \item \textbf{Normalization:} Normalize the translated keypoint \(\mathbf{p}'\) to a \([0, 1]\) range based on the image size:
    \begin{equation}
        \mathbf{p}_{\text{norm}} 
        = \Bigl( 
            \frac{\mathbf{p'_x}}{w_{\text{img}}}, 
            \frac{\mathbf{p'_y}}{h_{\text{img}}} 
        \Bigr).
    \end{equation}

    \item \textbf{Scaling:} Scale the normalized keypoint to the environment size to obtain global coordinates:
    \begin{equation}
        \mathbf{p}_{\text{global}} 
        = \Bigl( 
            \mathbf{p}_{\text{norm}, x} \times w_{\text{env}}, 
            \; \mathbf{p}_{\text{norm}, y} \times h_{\text{env}} 
        \Bigr).
    \end{equation}
\end{enumerate}

Applying this transformation across all keypoint pairs allows us to determine their positions within the global coordinate system, thereby accurately reflecting both the object's location in the image and the shift of the overall environment's origin.

\section*{C. Integrating Height Information into Keypoint Pairs}

Integrating height information into keypoint pairs is essential for accurately representing three-dimensional positions required for different tasks. 
Depending on the specific task, there are three ways of integrating the object's height into the keypoint pairs:

\begin{enumerate}
    \item both keypoints integrate object's height,
    \item only starting keypoint integrates object's height, and
    \item only ending keypoint integrates object's height.
\end{enumerate}

For example, the height integration on the cutting task involves starting at the height of the object's top surface with some safety margin and ending at the height of its base, such as the cutting board's surface. This pattern signifies a downward cutting motion from the top of the object to the base.

By adjusting the height information in the keypoint pairs according to the task requirements, we represent the three-dimensional motion paths needed for executing the task.
This task-specific integration of height information ensures that the robot's motions are suitable for the intended interactions with objects in the environment.

\begin{figure}[H]
    \centering
    \begin{minipage}{0.225\textwidth}
        \includegraphics[width=\textwidth]{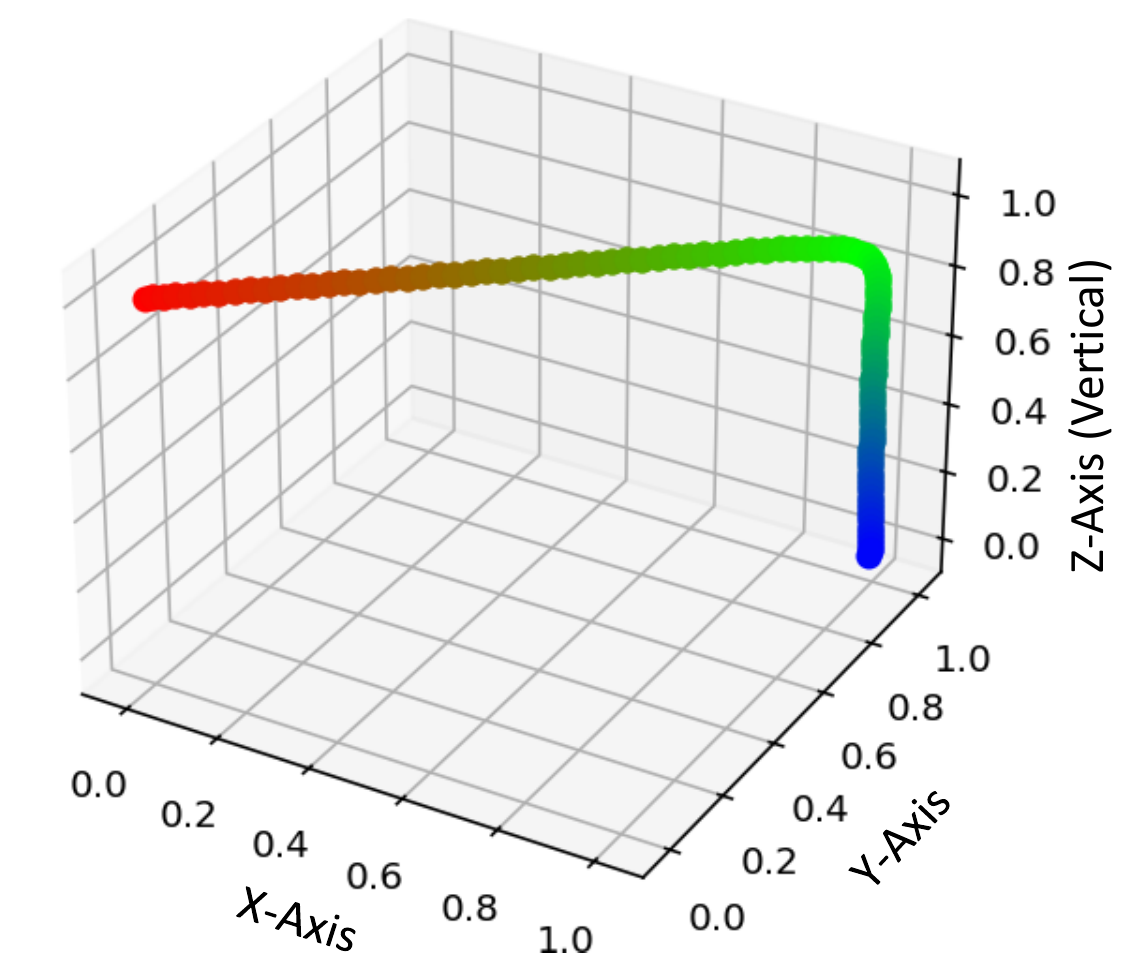}
        \centering
        \small (a) Straight-downward
    \end{minipage}
    \begin{minipage}{0.225\textwidth}
        \includegraphics[width=\textwidth]{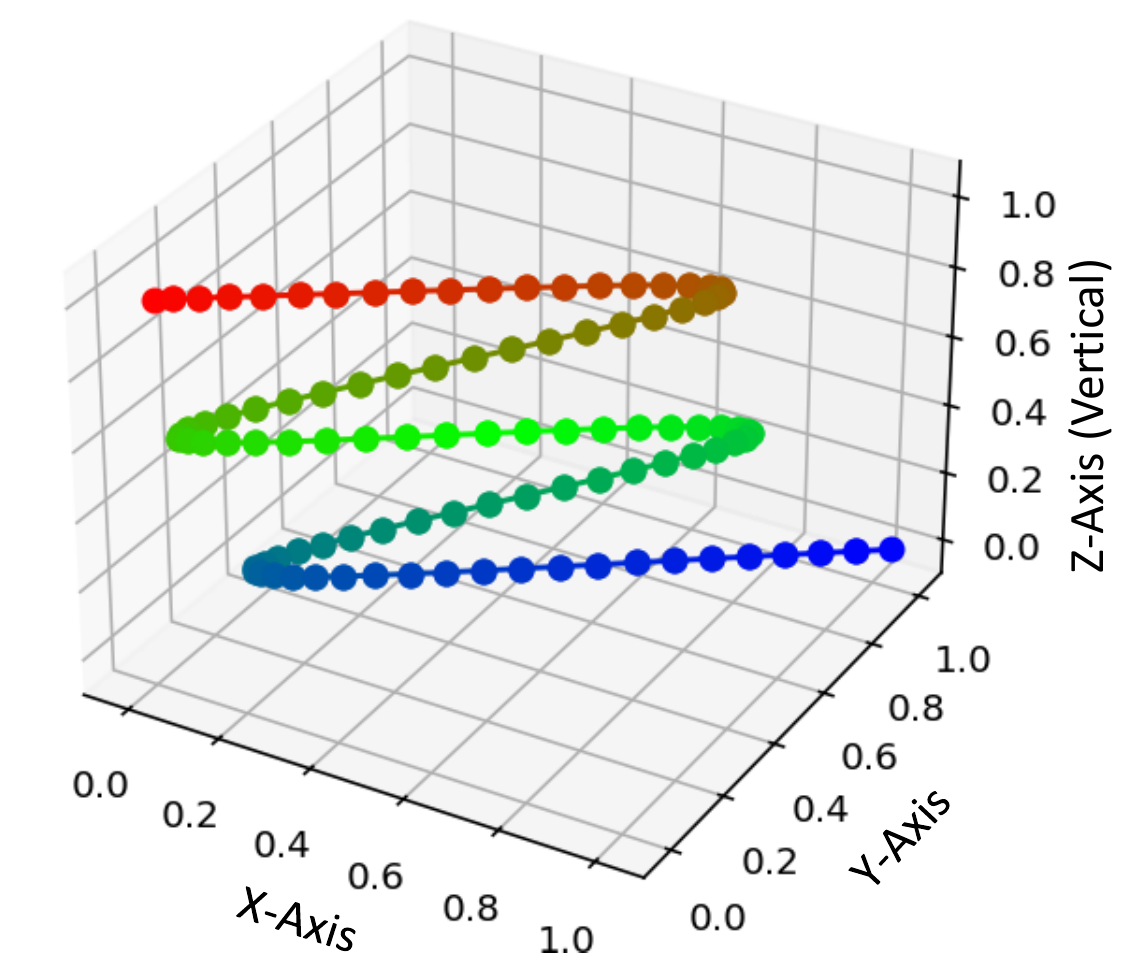}
        \centering
        \small (b) Sawing
    \end{minipage}
    \caption{3D projection of the object-cutting primitives used in simulation. 
    Time progression is represented by the change in color from red to green to blue.}
    \label{fig:sim_cutting_primitives_1}
\end{figure}

\begin{figure}[H]
    \centering
    \begin{minipage}{0.225\textwidth}
        \includegraphics[width=\textwidth]{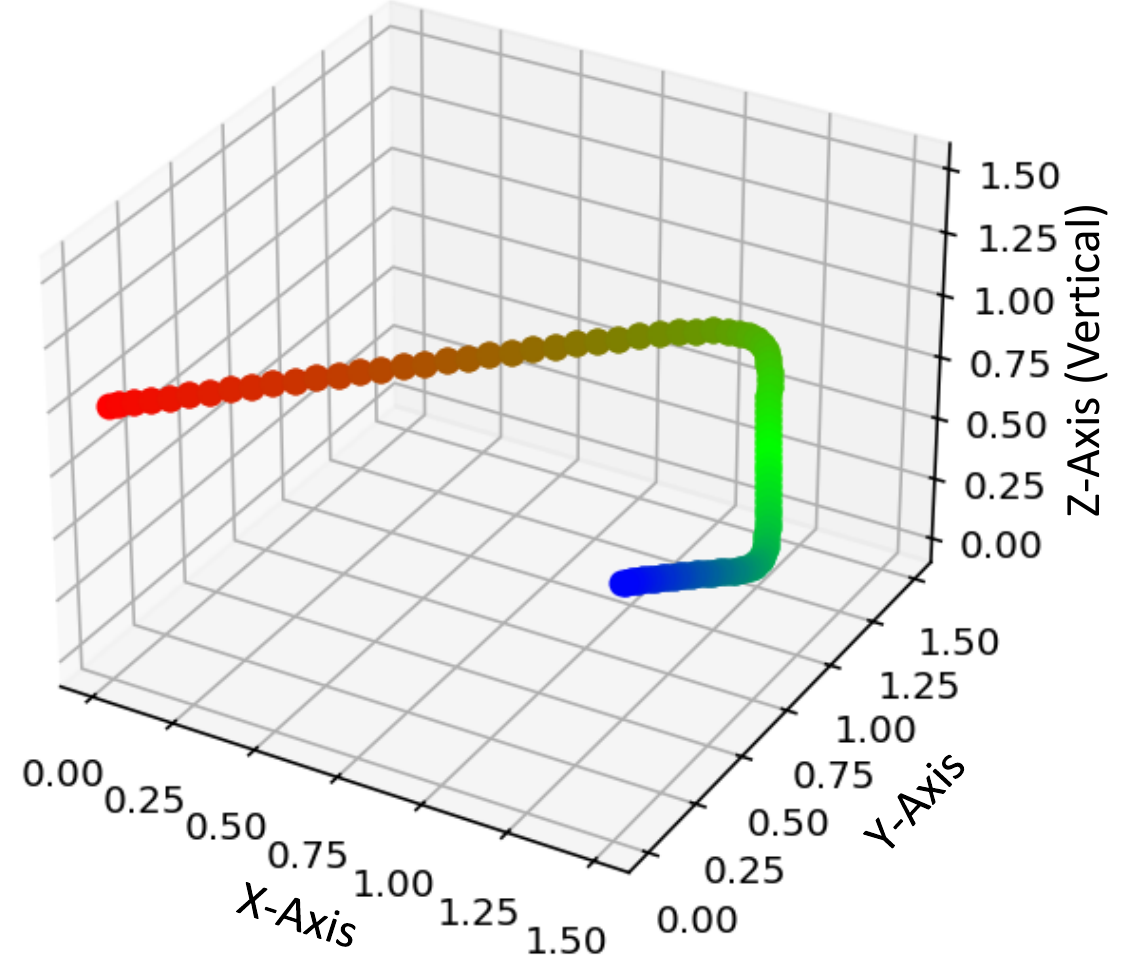}
        \centering
        \small (a) Downward
    \end{minipage}
    \begin{minipage}{0.225\textwidth}
        \includegraphics[width=\textwidth]{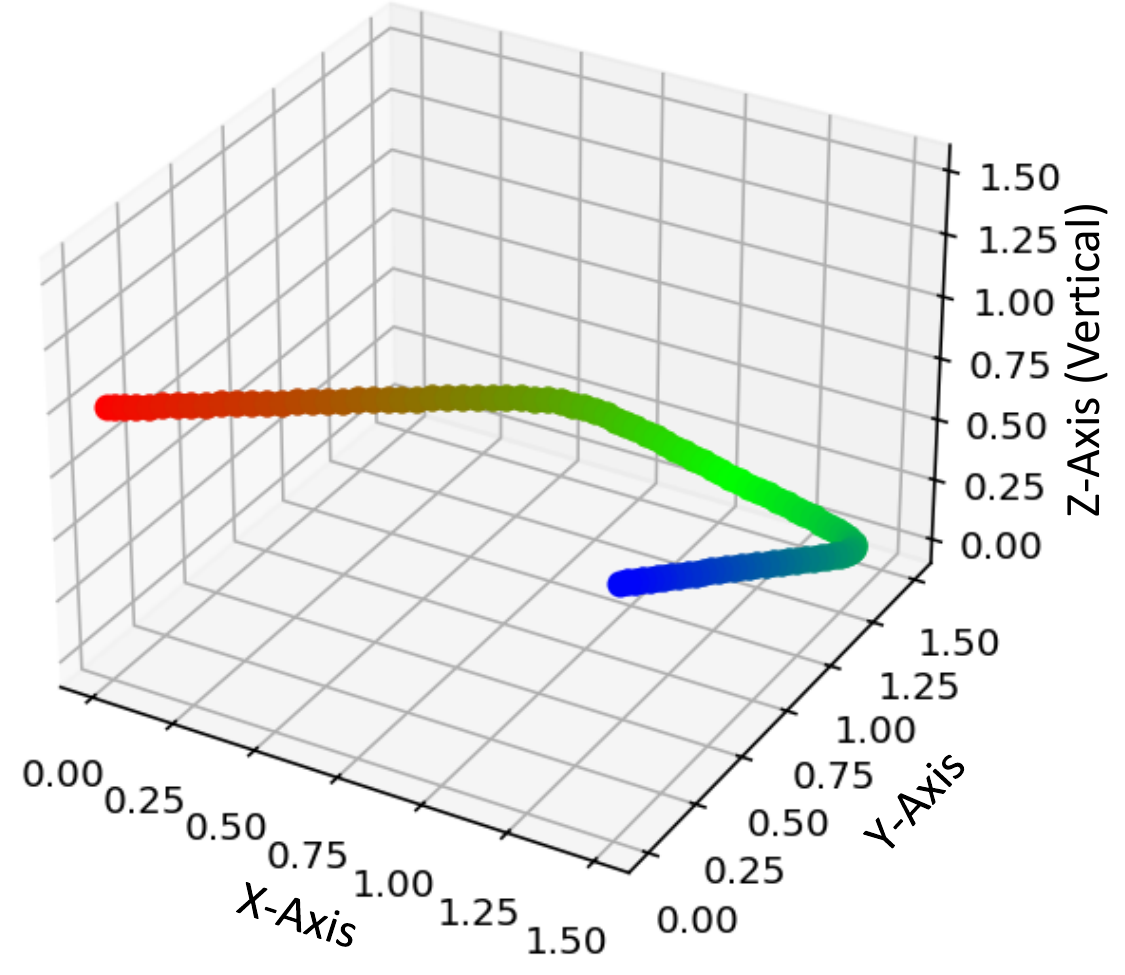}
        \centering
        \small (b) Forward
    \end{minipage}
    \caption{3D projection of the object cutting primitives used in a real robot environment. 
    Time progression is represented by the change in color from red to green to blue.}
    \label{fig:real_cutting_primitives_1}
\end{figure}

\begin{figure}[H]
    \centering
    \includegraphics[width=0.45\hsize]{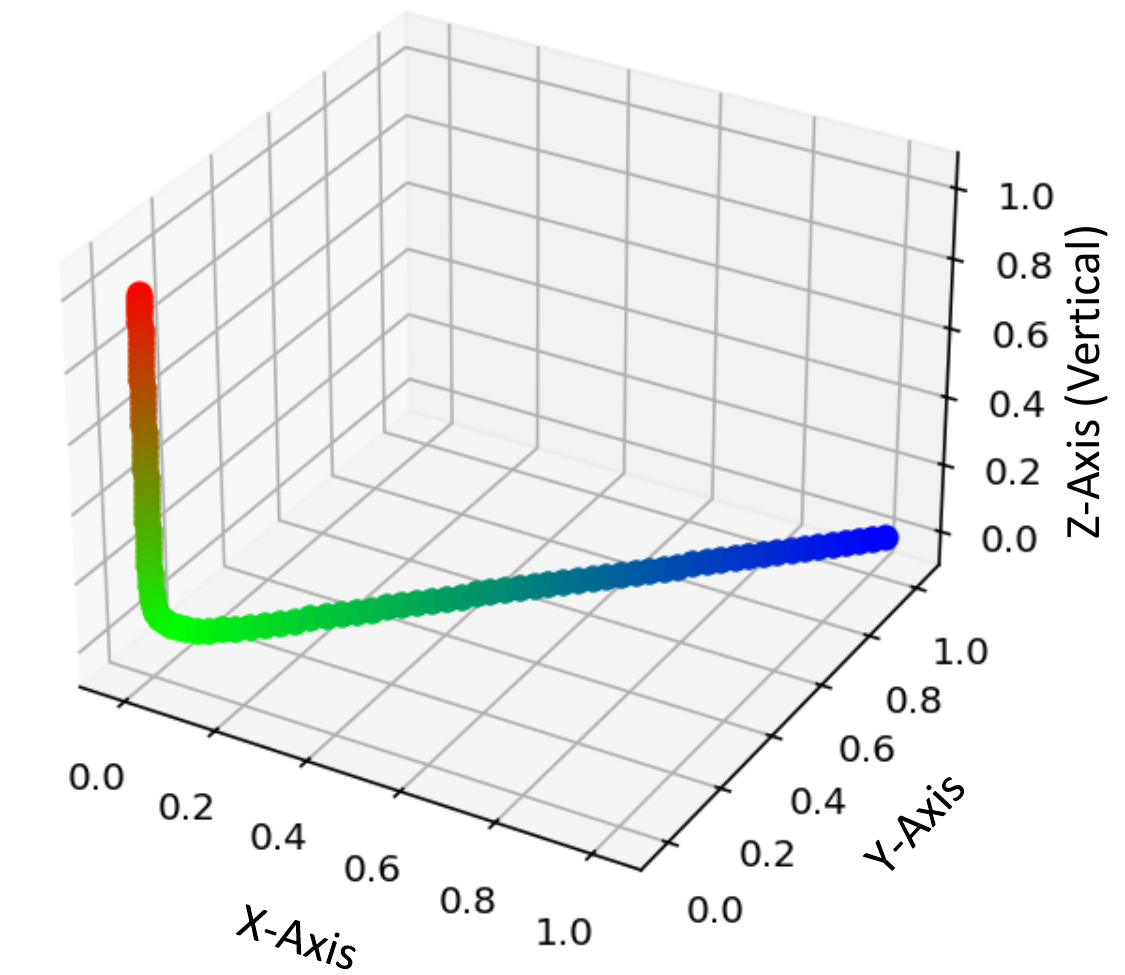}
    \caption{3D projection of the icing line primitive used in simulation.
    Time progression is represented by the change in color from red to green to blue.}
    \label{fig:sim_icing_primitives}
\end{figure}

\begin{figure*}[t]
\centering
\includegraphics[width=1\hsize]{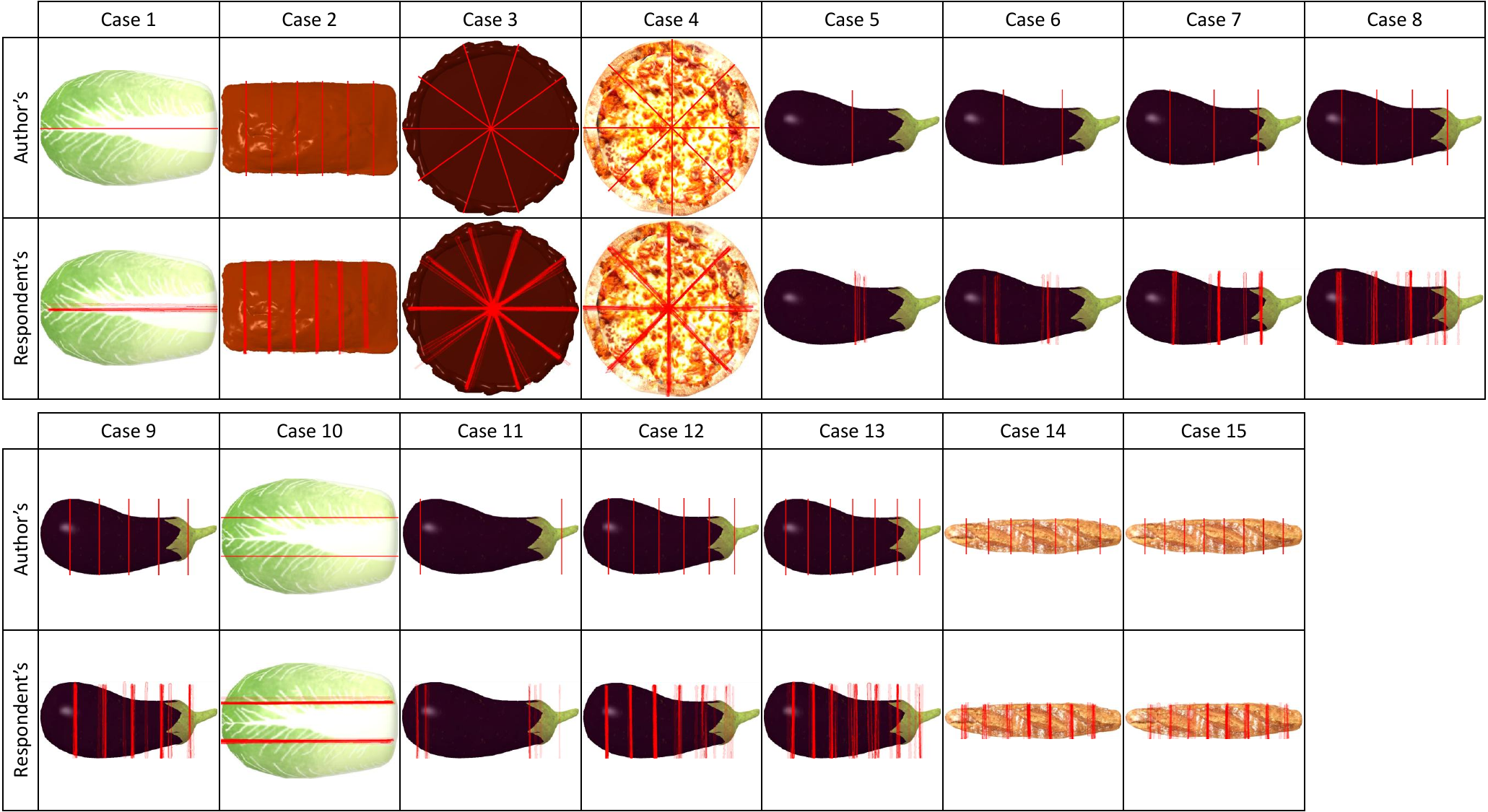}
\caption{Visualization of the user survey results conducted to validate our predefined ground-truth lines. The heatmap shows the aggregated drawings from five participants for various cutting tasks. The intensity of the red color indicates the frequency of drawn lines in that area, with stronger red signifying higher agreement among subjects. The analysis confirmed that our established ground-truth lines align well with these high-frequency areas.}
\label{fig:ground_truth_survey}
\end{figure*}

\section*{D. Visualizations of the Primitives Used in the Experiments}
\label{appendix:visualizations}

This appendix offers visual representations of the primitives used in our experiments, designed to facilitate specific actions within the simulation and real-world robot tasks. 
Visualizations of the primitives for the simulation experiments are presented in \figref{fig:sim_cutting_primitives_1} for the cutting task and \figref{fig:sim_icing_primitives} for the icing task, while the primitive used in the real robot environment is presented in \figref{fig:real_cutting_primitives_1}.

\section*{E. Object Cutting Task Ground Truth Lines Survey Results}
\label{appendix:ground-truth-survey}

To confirm that the ground-truth lines we established from domain knowledge were consistent with general user expectations, we conducted a validation survey with five consenting subjects (ages: 24-28). 
The survey procedure for each object cutting task case, as detailed in \tableref{table:cutting-tasks}, consisted of providing participants with an image of the object, an indicator of its real-world size, and the corresponding text prompt.
The subjects were then instructed to draw the cutting lines they would expect a robot to perform directly onto the image using a marker.

The collected drawings were aggregated and visualized as a heatmap in \figref{fig:ground_truth_survey}. 
In this visualization, the intensity of the red color corresponds to the frequency with which participants drew a line in a specific area, indicating spatial agreement. 
This analysis confirmed that our pre-established ground-truth lines are consistent with the most commonly agreed-upon cutting patterns identified in the survey, validating our initial expert-based approach.

\section*{F. Ablation: Necessity of Image Input}
\label{sim-exp5}

\subsection{Objective}

We conducted an ablation study to assess the necessity of image input by removing visual grounding from VLMs and evaluating their performance on generating spatially accurate keypoint designs in robotic cutting tasks. 
This comparison against text-only approaches helped determine if visual information is essential for resolving ambiguities in language descriptions, such as assumptions about object shape, to achieve geometrically valid keypoint pair designs.

For each approach, we generated 50 keypoint pair designs per cake shape (round/square) and measured the success rate on the basis of the methods' ability to divide the cake into six equal parts.
This quantifies the necessity of visual input for spatial precision in tasks where object geometry cannot be uniquely inferred from text.

\subsection{Comparison Methods}

\begin{table}[t]
\caption{\textbf{Comparison of input modalities for each method in the ablation study on the importance of image input}}
\label{table:input-info}
\setlength{\tabcolsep}{3pt}
\begin{tabular}{|l|>{\centering\arraybackslash}p{1.9cm}|>{\centering\arraybackslash}p{1.9cm}|>{\centering\arraybackslash}p{1.9cm}|}
\hline
Method & Object Info (Text) & Shape Info (Text) & Image Input \\
\hline
KeyMPs (ours) & \checkmark & - & \checkmark \\
KeyMPs-text & \checkmark & \checkmark & - \\
\hline
\end{tabular}
\end{table}

We evaluated our framework against ablation methods that employed LLMs instead of VLMs, referred to as \textbf{KeyMPs-text}. 
The prompts for \textbf{KeyMPs-text} depended only on linguistic input containing both the name of the object together with its shape. 
\tableref{table:input-info} summarizes the information received by each method.

\subsection{Results}

\begin{figure*}[t]
\centering
\includegraphics[width=\hsize]{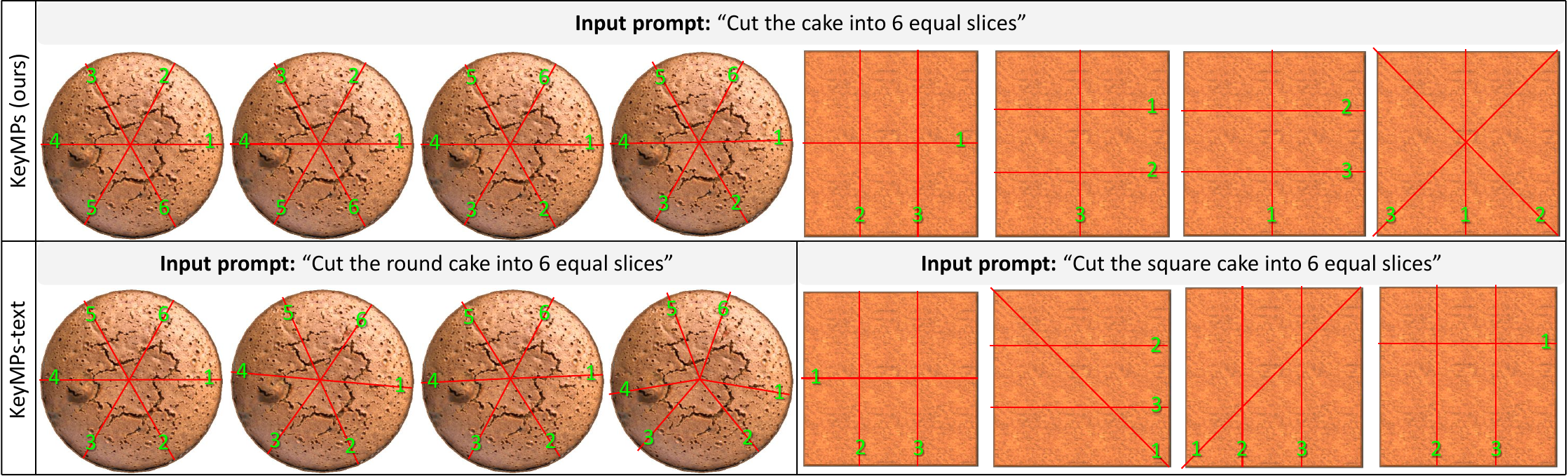}
\caption{Eight keypoint pair design samples generated by each method for cutting either a round or square cake into six equal slices. 
KeyMPs-text is provided with the additional context of shape (round/square) in the input prompt.
}
\label{fig:ex_3}
\end{figure*}

As illustrated in \figref{fig:ex_3}, without image input, the LLMs often relied on prior knowledge or assumptions about the object's shape, which might not align with the actual task requirements. 
This limitation is evident in the results of \textbf{KeyMPs-text}, where the generated designs are less consistent compared with those produced by \textbf{KeyMPs}.

The success rates, summarized in Table~\ref{table:exp3-results}, demonstrate that \textbf{KeyMPs} achieved significantly higher success rates, particularly with near-perfect accuracy (\SI{100}{\percent}) for the round cake. 
In contrast, \textbf{KeyMPs-text} attained only a \SI{34}{\percent} success rate, a substantial performance gap. 
Even when explicit textual descriptions of the object’s shape were provided, \textbf{KeyMPs-text} still underperformed against \textbf{KeyMPs}.

These findings underscore the importance of visual grounding in VLMs for spatial reasoning tasks: while the language-based method (\textbf{KeyMPs-text}) relies on error-prone prior assumptions about object geometry, the multimodal approach of \textbf{KeyMPs} leverages direct visual perception of shape and scale. 
This is particularly important in tasks involving objects with varying or non-unique shapes, such as slicing cakes with different geometries (e.g., round vs. square). 
For example, \textbf{KeyMPs} successfully handled round cakes with \SI{100}{\percent} accuracy and had a reliable level of performance on square cakes (\SI{64}{\percent}), whereas \textbf{KeyMPs-text} was significantly less effective and only partially recovered with explicit shape descriptions. 
These results demonstrate that integrating visual input is essential for precise keypoint design generation in real-world robotic applications.

\begin{table}[t]
\caption{\textbf{Success rates for keypoint pair generation in the image input ablation study}}
\label{table:exp3-results}
\setlength{\tabcolsep}{3pt}
\begin{tabular}{|l|>{\centering\arraybackslash}p{3cm}|>{\centering\arraybackslash}p{3cm}|}
\hline
Method & Round Cake & Square Cake \\
\hline
KeyMPs (ours) & 100\% & 64\% \\
KeyMPs-text & 78\% & 34\% \\
\hline
\end{tabular}
\end{table}


\end{document}